\documentclass[aps,prx,reprint,superscriptaddress,longbibliography,nofootinbib,floatfix]{revtex4-2}

\usepackage{amsmath,amssymb,amsfonts}
\usepackage[utf8]{inputenc}
\usepackage{array}
\usepackage{booktabs}
\usepackage{graphicx}
\usepackage{xcolor}
\usepackage{hyperref}
\hypersetup{
    colorlinks=true,
    linkcolor=blue,
    filecolor=magenta,
    urlcolor=cyan,
    pdftitle={ProvenanceGuard},
    pdfpagemode=FullScreen,
}
\usepackage{siunitx}
\usepackage{textcomp}
\usepackage{threeparttable}
\usepackage{url}
\usepackage{multirow}
\usepackage{verbatim}
\usepackage{tikz}
\usepackage{placeins}

\usetikzlibrary{arrows.meta,positioning,shapes.geometric,fit,calc}
\newcommand{\ours}{\textsc{ProvenanceGuard}}
\newcommand{\adjlabels}{LLM-assisted}

\newcommand{\artifact}[1]{\nolinkurl{#1}}
\newcommand{\hashartifact}[1]{\nolinkurl{#1}}

\newcommand{\tablefontsize}{\footnotesize}
\definecolor{oursblue}{HTML}{1F77B4}
\definecolor{baselinegray}{HTML}{7F7F7F}
\definecolor{accentorange}{HTML}{FF7F0E}
\definecolor{accentgreen}{HTML}{2CA02C}
\definecolor{accentred}{HTML}{D62728}

\makeatletter
\def\frontmatter@title@format{\centering\LARGE\bfseries\let\thanks\thanks@latex}
\makeatother

\begin{document}

\makeatletter
\immediate\write\@auxout{\string\citation{provenanceguardBibControl}}
\makeatother

\title{ProvenanceGuard: Source-Aware Factuality Verification for MCP-Based LLM Agents}

\author{Ander Alvarez}
\email[Email: ]{ander.alvarez@multiversecomputing.com}
\affiliation{Multiverse Computing, Parque Cientifico y Tecnológico de Gipuzkoa, Paseo de Miramón, 170, 20014 Donostia / San Sebastián, Spain}

\author{Santhiya Rajan}
\email[Email: ]{santhiya.rajan@multiversecomputing.com}
\affiliation{Multiverse Computing, Parque Cientifico y Tecnológico de Gipuzkoa, Paseo de Miramón, 170, 20014 Donostia / San Sebastián, Spain}

\author{Alessandro Genuardi}
\email[Email: ]{alessandro.genuardi@multiversecomputing.com}
\affiliation{Multiverse Computing, Parque Cientifico y Tecnológico de Gipuzkoa, Paseo de Miramón, 170, 20014 Donostia / San Sebastián, Spain}

\author{Oliver Wirjadi}
\email[Email: ]{oliver.wirjadi@multiversecomputing.com}
\affiliation{Multiverse Computing, Parque Cientifico y Tecnológico de Gipuzkoa, Paseo de Miramón, 170, 20014 Donostia / San Sebastián, Spain}

\author{Samuel Mugel}
\email[Email: ]{sam.mugel@multiversecomputing.com}
\affiliation{Multiverse Computing, Centre for Social Innovation, 192 Spadina Avenue Suite 509, Toronto, ON M5T 2C2, Canada}

\author{Rom\'an Or\'us}
\email[Email: ]{roman.orus@multiversecomputing.com}
\affiliation{Multiverse Computing, Parque Cientifico y Tecnológico de Gipuzkoa, Paseo de Miramón, 170, 20014 Donostia / San Sebastián, Spain}
\affiliation{Donostia International Physics Center, Paseo Manuel de Lardizabal 4, E-20018 San Sebastián, Spain}
\affiliation{Ikerbasque Foundation for Science, Maria Diaz de Haro 3, E-48013 Bilbao, Spain}

\begin{abstract}
Tool-using LLM agents increasingly produce answers from heterogeneous evidence sources exposed through Model Context Protocol (MCP) servers, including search results, APIs, databases, records, formulary tools, and other external systems. Existing factuality and faithfulness metrics typically evaluate whether an answer is supported by the available context after evidence has been pooled. This abstraction misses an important provenance-sensitive failure mode: a claim may be supported somewhere in the evidence while being attributed to the wrong source. We call this failure cross-source conflation.

We introduce \ours{}, a source-aware verifier for MCP-grounded answers. \ours{} is a verification layer with calibrated router and natural-language-inference (NLI) system: it consumes captured MCP traces with stable tool IDs, source IDs, and raw tool outputs; decomposes an answer into atomic claims; routes claims to source-specific evidence; checks support with NLI and an attention-derived token-alignment proxy; and separately compares the claim's stated attribution with the routed source. It returns both per-claim source verdicts and an answer-level allow/block decision, and can invoke retrieval-augmented answer revision (RARR)-style repair to revise blocked answers before re-verification.

We instantiate this verifier on a frozen captured corpus of 281 real medical MCP-agent traces, using the medical agent as a concrete testbed rather than as a restriction of the framework. A 266-trace claim-adjudicated subset yields 2,325 \adjlabels{} claim labels, split by trace into train, validation, and held-out test partitions; the 361 held-out labels are then verified by human experts, and the full 281-trace corpus is used for answer-level repair evaluation. On the 40-trace, 361-claim held-out split, \ours{} reaches reject/block F1 0.802 and source accuracy 0.858 over 260 source-eligible claims. Source-blind claim/evidence baselines on the same packet reach reject/block F1 0.783 for MiniCheck, 0.758 for RAGAS Faithfulness, 0.662 for AlignScore, and 0.436 for SummaC-ZS, but none emit claim-to-source IDs. On a harder multi-source adjudicated benchmark, \ours{} reaches reject/block F1 0.846 over frozen extracted claims from the locked test questions, while source-plus-relation accuracy drops to 0.229, showing that exact source ownership remains difficult under many semantically close candidate sources. On full real traces, a repair-and-reverify loop resolves all 173 blocked answers, though 144 require fallback text rather than substantive rewriting; on reconstructed multi-source test traces, a fresh answer-level repair rerun resolves all 59 initially blocked answers with two terminal fallbacks. In 50 targeted source-conflation probes over frozen MCP evidence, \ours{} detects all 50 deliberately injected source-attribution swaps in controlled probes with no retained wrong attribution. Operationally, \ours{} acts as a conservative post-generation gate: it removes nearly all held-out claims that should not pass, but it also sends some supported claims to review or repair.

\end{abstract}


\maketitle

\section{Introduction}

LLM agents that use Model Context Protocol (MCP) increasingly operate over multiple external tools rather than a single retrieved passage. They call tools through MCP servers, inspect structured records, combine multiple outputs, and often generate answers that mix source-grounded facts with general background knowledge and safety disclaimers. The general verification problem is source ownership: factuality verification must assess not only whether a claim is supported somewhere in the available evidence, but also whether the answer assigns the claim to the correct source. This problem arises whenever an answer combines records, policy documents, search results, tickets, databases, or API outputs that carry distinct provenance.

Consider an answer from a customer support agent that states: ``According to the account record, this plan includes a 30-day refund window.'' The refund-window claim may be supported by a policy document, but not by the account record. A source-blind verifier that pools the account record and the policy document can mark the claim as supported. A source-aware verifier should instead reject this attribution: the claim may be true in one source, but the answer assigns it to the wrong source.

The same issue appears in a medical agent: a patient-specific medication fact may be supported by a patient-history tool output but become misleading if the answer presents it as a literature finding. We use this medical agent only as the empirical testbed for demonstration; the verification problem is the general source-ownership problem above for all domains.

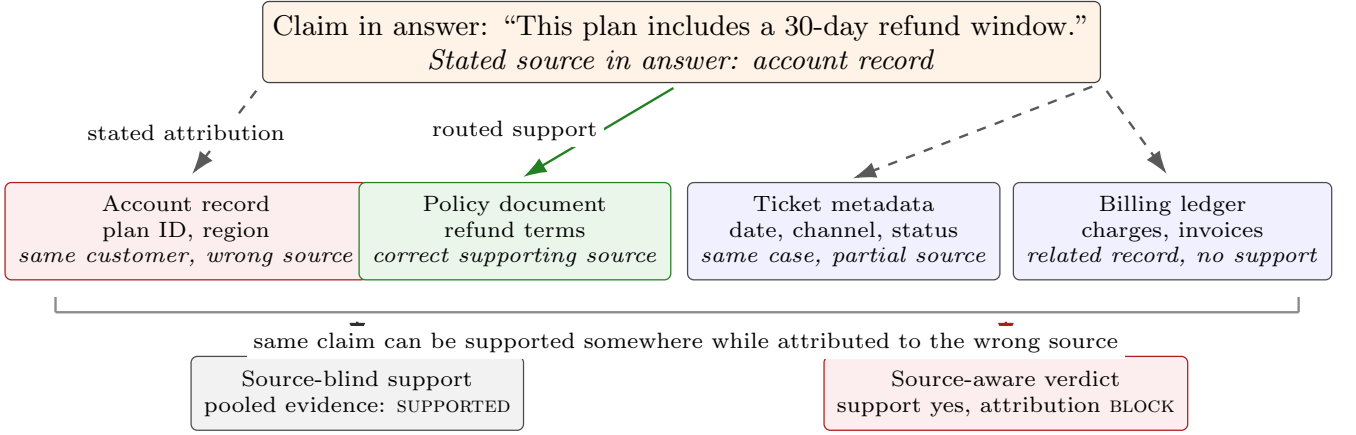
\begin{figure*}[t]
  \centering
  \resizebox{0.98\textwidth}{!}{%
\begin{tikzpicture}[
  x=1cm,
  y=1cm,
  claim/.style={draw=black!75, rounded corners=2pt, align=center, fill=orange!10, minimum width=7.2cm, minimum height=0.90cm, font=\small},
  source/.style={draw=black!70, rounded corners=2pt, align=center, fill=blue!6, minimum width=3.45cm, minimum height=1.08cm, inner sep=4pt, font=\scriptsize},
  correct/.style={draw=accentgreen!80!black, rounded corners=2pt, align=center, fill=accentgreen!10, minimum width=3.45cm, minimum height=1.08cm, inner sep=4pt, font=\scriptsize},
  wrong/.style={draw=accentred!80!black, rounded corners=2pt, align=center, fill=accentred!8, minimum width=3.45cm, minimum height=1.08cm, inner sep=4pt, font=\scriptsize},
  verdict/.style={draw=black!70, rounded corners=2pt, align=center, minimum width=3.45cm, minimum height=0.82cm, inner sep=4pt, font=\scriptsize},
  arrow/.style={-{Latex[length=2.6mm,width=1.8mm]}, line width=0.75pt, draw=black!82, shorten >=3pt, shorten <=3pt},
  dasharrow/.style={-{Latex[length=2.6mm,width=1.8mm]}, line width=0.75pt, dashed, draw=black!65, shorten >=3pt, shorten <=3pt},
  bracket/.style={line width=0.75pt, draw=black!45},
  label/.style={font=\scriptsize, fill=white, inner sep=2pt}
]

\node[claim] (claim) at (7.2,4.45) {Claim in answer: ``This plan includes a 30-day refund window.''\\
\textit{Stated source in answer: account record}};

\node[wrong] (chart) at (1.7,2.35) {Account record\\plan ID, region\\\textit{same customer, wrong source}};
\node[correct] (trial) at (5.35,2.35) {Policy document\\refund terms\\\textit{correct supporting source}};
\node[source] (metadata) at (9.00,2.35) {Ticket metadata\\date, channel, status\\\textit{same case, partial source}};
\node[source] (labs) at (12.65,2.35) {Billing ledger\\charges, invoices\\\textit{related record, no support}};

\node[verdict, fill=gray!10] (pooled) at (3.60,0.55) {Source-blind support\\pooled evidence: \textsc{supported}};
\node[verdict, fill=accentred!8, draw=accentred!80!black] (aware) at (10.80,0.55) {Source-aware verdict\\support yes, attribution \textsc{block}};

\draw[dasharrow] (claim.south west) -- (chart.north);
\draw[arrow, draw=accentgreen!80!black] (claim.south) -- (trial.north);
\draw[dasharrow] (claim.south east) -- (metadata.north);
\draw[dasharrow] (claim.south east) -- (labs.north);

\draw[bracket] (0.25,1.45) -- (14.05,1.45);
\draw[bracket] (0.25,1.45) -- (0.25,1.62);
\draw[bracket] (14.05,1.45) -- (14.05,1.62);
\draw[arrow] (3.60,1.45) -- (pooled.north);
\draw[arrow, draw=accentred!80!black] (10.80,1.45) -- (aware.north);

\node[label, anchor=south] at (5.35,3.28) {routed support};
\node[label, anchor=south] at (1.7,3.28) {stated attribution};
\node[label, anchor=north] at (7.25,1.30) {same claim can be supported somewhere while attributed to the wrong source};

\end{tikzpicture}%
}
  \caption{Why source-aware factuality is stricter than source-blind support. A claim can be supported by one MCP source while the answer attributes it to another. Source-blind scoring sees support in pooled evidence; \ours{} separately checks whether the supporting source matches the stated or implied attribution.}
  \label{fig:source_ambiguity}
\end{figure*}

Figure~\ref{fig:source_ambiguity} summarizes the difference between pooled support and source-aware support: the same claim can be true relative to one source while being incorrectly attributed to another.

This distinction matters because many factuality systems were designed for summary consistency or retrieval faithfulness. They typically score support against a context, rather than claim-to-source ownership. Such scores are informative for unsupported fabrication, but they are insufficient for MCP-grounded agents whose answers carry implicit or explicit provenance claims.

This paper makes four contributions:

\begin{enumerate}
  \item We formulate source-attribution factuality for MCP-grounded answers: a verifier must evaluate both support and source ownership for each claim.
  \item We introduce \ours{}, a calibrated source-aware Router+NLI verifier that decomposes answers into claims, preserves stable MCP tool IDs and source IDs from raw tool outputs, routes claims to source-specific evidence, and detects cross-source conflation.
  \item We define reporting axes for source-aware verification that separate binary support gating from exact verdict typing, source ownership, source-plus-relation correctness, and repair outcomes.
  \item We characterize the practical deployment tradeoff of a fail-closed verifier: how many unsupported claims it removes, how many supported claims it sends to review or repair, what source attribution signal it adds, and what latency overhead it introduces.
\end{enumerate}

To evaluate the system concretely, we instantiate the trace interface on a medical MCP agent. This use case is introduced as an empirical testbed, not as part of the definition of \ours{}. It is useful because its tool outputs include multiple provenance-bearing source families with expert-verifiable claims; the domain-specific tools and source families are described in Section~\ref{sec:setup}. The 281-trace real-agent corpus does not by itself estimate the natural prevalence of cross-source conflation: the captured random traces contain single source-family outputs, or search plus metadata outputs that remain within the same broad evidence family. We therefore evaluate cross-source conflation with a targeted 50-case source-confusion benchmark over frozen MCP evidence, where each probe contains one deliberate attribution swap.

The central claim is limited to source-attribution factuality in MCP-grounded answers; we do not claim to solve open-domain factuality detection, domain safety validation, or parametric-knowledge correction.

\section{Related Work}

Our work sits at the intersection of support verification, source attribution, and tool-grounded agent evaluation. We organize prior work by the kind of evidence object each line of work preserves.

\paragraph{Fine-grained support verification.}
Factual consistency systems such as FActScore, MiniCheck, SummaC, AlignScore, and VeriScore evaluate whether generated claims are supported by evidence \citep{factscore,minicheck,summac,alignscore,veriscore}. Fine-grained work decomposes generated text below the sentence level: Dependency Arc Entailment localizes errors at dependency arcs \citep{goyal2020dae}, QASemConsistency expresses predicate-argument propositions as question-answer pairs \citep{cattan2024qasem}, and PrefixNLI studies entailment over generation prefixes \citep{harary2025prefixnli}. These methods motivate claim-level checking, but their standard outputs do not preserve stable claim-to-MCP-source IDs. They are therefore relevant to binary allow/block decisions but are not direct baselines for source attribution metrics such as Top-1 source accuracy, recall@k, mean reciprocal rank, or source-set Jaccard.

\paragraph{RAG faithfulness and attributed generation.}
RAG evaluation frameworks such as RAGAS ask whether answer statements are faithful to retrieved context \citep{ragas}. Other work studies answer generation with citations and attribution: ALCE evaluates citation quality for LLM-generated answers \citep{alce}; AttributedQA formalizes attributed question answering \citep{attributedqa}; AutoAIS automates AIS-style attribution judgments \citep{autoais,ais}; and TRUE consolidates factual-consistency datasets across summarization, dialogue, paraphrasing, and verification \citep{true}. More recent attribution work localizes evidence to user-selected spans through LAQuer \citep{hirsch2025laquer} or decomposes generation into executable attribution programs \citep{wan2025generationprograms}. 
These methods are close in motivation, but faithfulness to pooled or cited context is not equivalent to source ownership. A claim can be supported by one retrieved source while being falsely attributed to another. ALCE \citep{alce} evaluates whether LLM-generated citations point to the correct supporting passage, which is the closest existing task to source attribution. However, ALCE operates at the passage or chunk level within a single retrieved set, whereas MCP traces expose stable tool-level source IDs that require a routing step to identify which tool output is responsible for a given claim. Our work can therefore be seen as extending citation-style attribution to the tool-provenance layer.

\paragraph{Tool and source attribution in multi-source systems.}
As RAG systems become tool-using agents, attribution must track which tool output supplied the evidence. Atomic Information Flow models tool outputs, LLM calls, and final responses as flows of atomic information through an orchestration graph \citep{gao2026aif}. FaithfulRAG focuses on conflicts between retrieved evidence and parametric knowledge at the fact level \citep{zhang2025faithfulrag}, while the Generate-but-Verify line of work couples answer generation with faithfulness prediction \citep{filice2025generate}. Our setting differs because MCP traces expose stable tool and source identifiers. We do not infer latent information flow or resolve parametric-knowledge conflicts; we verify whether the answer's stated or implied attribution matches the routed MCP source.

\paragraph{Post-hoc revision and trained verifiers.}
RARR \citep{rarr} takes a generated passage, researches evidence, and revises unsupported claims while preserving the original style and structure. In our setting, RARR-style repair is evaluated after source-aware blocking: the verifier rejects an answer, repair attempts to produce a source-grounded revision or fallback text with no evidence-requiring factual claim, and the same verifier rechecks the revised answer. Training-based approaches improve consistency at generation or detection time, including reinforcement learning with textual entailment feedback \citep{rlef}, FactCC's synthetic inconsistency classifier \citep{factcc}, and RAGulator's lightweight out-of-context detectors for grounded generation \citep{poey2025ragulator}.
These methods are complementary to ProvenanceGuard, which operates as an independent post-hoc check on black-box MCP-agent outputs. For calibration specifically, alternatives include Platt scaling, isotonic regression, and conformal prediction; we adopt a random-forest calibrator because it handles tabular verifier features after simple one-hot encoding and numeric scaling.

\paragraph{Tool-use and agent traces.}
Tool-use evaluation often studies whether agents call the right tools, complete tasks, and produce valid final answers. Our focus is narrower: after an agent has produced an answer and the tool trace is available, can a verifier decide whether each claim is supported by the source the answer cites or implies? The MCP community has independently identified this gap: the proposed verification capability for MCP servers \citep{mcpverification} calls for structured verdicts with confidence scores, mirroring our allow/block/unavailable decision space. Our work provides a concrete implementation of this verification layer for the MCP trace setting.

\section{\ours{}}
\label{sec:method}

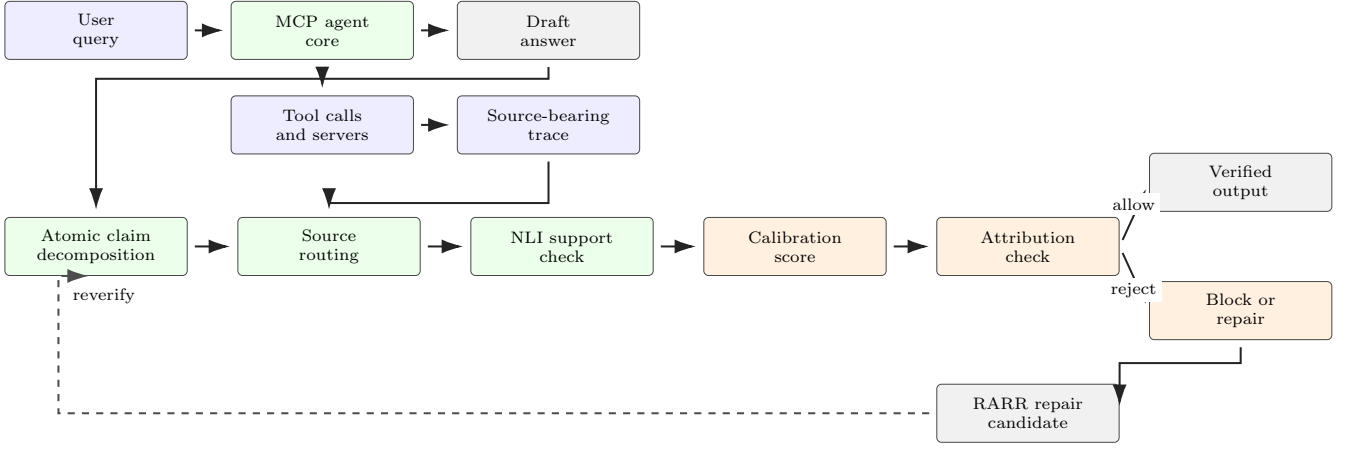
\begin{figure*}[t]
  \centering
  \resizebox{0.98\textwidth}{!}{%
\begin{tikzpicture}[
  x=1cm,
  y=1cm,
  box/.style={
    draw=black!72,
    rounded corners=2pt,
    align=center,
    minimum width=2.70cm,
    minimum height=0.86cm,
    inner sep=4pt,
    font=\scriptsize
  },
  source/.style={box, fill=blue!7},
  process/.style={box, fill=green!8},
  decision/.style={box, fill=orange!12},
  output/.style={box, fill=gray!11},
  group/.style={draw=black!42, rounded corners=3pt, inner xsep=12pt, inner ysep=12pt},
  arrow/.style={-{Latex[length=3.2mm,width=2.1mm]}, line width=0.85pt, draw=black!86, shorten >=3pt, shorten <=3pt},
  feedback/.style={-{Latex[length=3.2mm,width=2.1mm]}, line width=0.85pt, draw=black!70, dashed, shorten >=3pt, shorten <=3pt},
  labeltext/.style={font=\scriptsize, fill=white, inner sep=2pt}
]

\node[source] (query) at (0,5.25) {User\\query};
\node[process] (agent) at (3.35,5.25) {MCP agent\\core};
\node[output] (draft) at (6.70,5.25) {Draft\\answer};

\node[source] (tools) at (3.35,3.85) {Tool calls\\and servers};
\node[source] (trace) at (6.70,3.85) {Source-bearing\\trace};

\node[process] (claims) at (0,2.05) {Atomic claim\\decomposition};
\node[process] (route) at (3.45,2.05) {Source\\routing};
\node[process] (nli) at (6.90,2.05) {NLI support\\check};
\node[decision] (calib) at (10.35,2.05) {Calibration\\score};
\node[decision] (attrib) at (13.80,2.05) {Attribution\\check};
\node[output] (allow) at (16.95,3.00) {Verified\\output};
\node[decision] (block) at (16.95,1.10) {Block or\\repair};
\node[output] (repair) at (13.80,-0.42) {RARR repair\\candidate};

\draw[arrow] (query) -- (agent);
\draw[arrow] (agent) -- (draft);
\draw[arrow] (agent.south) -- (tools.north);
\draw[arrow] (tools) -- (trace);

\draw[arrow] (draft.south) -- ++(0,-0.28) -| (claims.north);
\draw[arrow] (trace.south) -- ++(0,-0.72) -| (route.north);

\draw[arrow] (claims) -- (route);
\draw[arrow] (route) -- (nli);
\draw[arrow] (nli) -- (calib);
\draw[arrow] (calib) -- (attrib);
\draw[arrow] (attrib.east) -- node[labeltext, above, pos=0.48] {allow} (allow.west);
\draw[arrow] (attrib.east) -- node[labeltext, below, pos=0.48] {reject} (block.west);
\draw[arrow] (block.south) -- ++(0,-0.34) -| (repair.east);
\draw[feedback] (repair.west) -- ++(-13.00,0) |- (claims.south);
\node[labeltext, anchor=south west] at ($(claims.south)+(-0.42,-0.48)$) {reverify};

\end{tikzpicture}%
}
  \caption{Sequential source-aware verification pipeline. The agent core calls MCP tools and produces a draft answer; \ours{} consumes the answer and captured source-bearing tool traces, decomposes the answer into claims, routes claims to MCP evidence, estimates support with natural-language-inference (NLI), token alignment, and calibration, checks attribution separately from support, and sends blocked answers through repair and re-verification.}
  \label{fig:pipeline}
\end{figure*}

\ours{} verifies attribution in provenance-preserving Model Context Protocol (MCP) traces. The method is not specific to any one agent or application domain: it assumes only that the trace preserves tool outputs and source identifiers. The goal is not to prove that a source is correct in every domain-specific sense. The goal is narrower: when an agent answer makes a factual claim, \ours{} asks whether the claim is supported by the source it should be attributed to, and whether the answer assigns the claim to the right MCP provenance object. This framing is important in data-sensitive domains, where an offline verifier can spend more computation on reliable source attribution rather than optimizing for interactive latency.

Figure~\ref{fig:pipeline} shows the full verification flow. The key design choice is that source identity is carried through decomposition, routing, support scoring, attribution checking, and repair rather than being collapsed into a single pooled context.

\subsection{Trace Interface}

\ours{} starts from the trace object produced by a tool-using agent. A trace contains the user request, the final assistant answer, and a list of complete tool outputs. Each evidence object is represented as
\begin{equation}
e_i = (\mathrm{tool}_i, \mathrm{source}_i, \mathrm{text}_i),
\label{eq:evidence}
\end{equation}
where \(\mathrm{tool}_i\) identifies the tool type, \(\mathrm{source}_i\) identifies the provenance-bearing source within the trace, and \(\mathrm{text}_i\) is the observed tool output or structured sub-output.
Multiple calls to the same tool type, such as two search calls with different queries, produce separate evidence objects if their source IDs differ. The verifier never collapses these evidence objects into a single anonymous context. It carries the source identifier forward because the same answer may combine record evidence, search results, database outputs, and metadata evidence.

The output of this stage is a source-preserving evidence set \(E=\{e_1,\ldots,e_n\}\) paired with the assistant answer \(y\).
When a tool output lacks a stable source ID, the verifier falls back to tool name as the provenance identifier. Appendix~\ref{app:trace_json} gives the escaped JSON trace structure used for this interface.

\subsection{Atomic Claims}

The next step decomposes the answer \(y\) into checkable claims,
\begin{equation}
C(y)=\{c_1,\ldots,c_m\}.
\label{eq:claimset}
\end{equation}
Here \(C(y)\) is the claim set extracted from answer \(y\), \(c_j\) is the \(j\)th claim, and \(m\) is the number of claims extracted for that answer. Each \(c_j\) should express one factual proposition. Literal values that can change the meaning of a claim, such as numbers, units, dates, identifiers, quantities, and quoted values, are preserved exactly. If the answer names a source family, such as an account record, database, search result, or policy document, the decomposer also preserves that stated attribution span.

Template disclaimers, such as ``consult a professional,'' are not treated as evidence-bearing content unless they contain a concrete factual assertion. This prevents template text from dominating the claim set while keeping the verifier accountable for factual statements that remain in the answer.

\subsection{Source Routing}

For each claim, \ours{} ranks the source-specific evidence objects before support checking. Let \(q_j\) be the embedding of claim \(c_j\), and let \(e_{s,1},\ldots,e_{s,n_s}\) be the embedding vectors for the \(n_s\) chunks belonging to source \(s\). The source representation is the centroid
\begin{equation}
r_s = \frac{1}{n_s}\sum_{i=1}^{n_s} e_{s,i}.
\label{eq:source_centroid}
\end{equation}
The router selects the highest-scoring source,
\begin{equation}
\hat{s}_j = \arg\max_s \cos(q_j,r_s),
\label{eq:routing}
\end{equation}
and records the routing margin between the top two source scores. The top-ranked source is used for single-source attribution. Top-\(k\) routing is used only as an analysis of whether the correct source is near the head of the ranked list.

After routing, the premise is narrowed to claim-relevant evidence and capped by a fixed length budget. This premise-selection step controls distractor effects before NLI scoring: shorter premises reduce irrelevant context, while longer premises may improve recall but increase truncation and distraction risk. The current service default is 512 tokens for the premise--claim pair, which avoids the earlier 256-token bottleneck while staying within the base DeBERTa context window.

\subsection{Support and Alignment}

The routed premise and claim are then passed to a natural-language-inference (NLI) model. The NLI relation is one of entailment, neutral, or contradiction. Entailment is evidence of support, contradiction is evidence of conflict, and neutral means the routed source does not provide enough support by itself.

\ours{} also computes a heuristic token-alignment proxy from NLI attention. This proxy is used as an additional grounding signal, not as an independently validated explanation of the NLI model. Let \(P\) be the premise token indices and \(H_c\) the claim token indices in the paired NLI encoding. For attention tensor \(A^{\ell,h}\), the alignment score for claim token \(i \in H_c\) is
\begin{equation}
a_i = \max_{j \in P}\left(\frac{1}{LK}\sum_{\ell=1}^{L}\sum_{h=1}^{K} A^{\ell,h}_{ij}\right),
\label{eq:alignment}
\end{equation}
where \(L\) is the number of layers and \(K\) is the number of attention heads. For a fixed claim token \(i\), this averages attention from \(i\) to each premise token \(j\) across all heads and layers, then keeps the strongest premise-token alignment. A content token is marked weakly grounded when
\begin{equation}
a_i < \tau \max_{k \in H_c} a_k,
\label{eq:grounding_threshold}
\end{equation}
where \(\tau\) is a fixed relative alignment threshold. The supported-token ratio is computed over non-stopword claim tokens. A claim with low token support remains blocked even if the coarse NLI label is uncertain.

Literal values receive a stricter check. If an entailed claim contains a protected value absent from the routed premise after normalization, the claim is treated as unsupported or contradictory. Conversely, a neutral NLI output can be lexically rescued only when all protected values are present and normalized content-term overlap is high. Let \(T_{\mathrm{claim}}\) and \(T_{\mathrm{premise}}\) be the sets of normalized non-stopword content terms in the claim and routed premise. The overlap score is
\begin{equation}
\frac{|T_{\mathrm{claim}}\cap T_{\mathrm{premise}}|}{|T_{\mathrm{claim}}|},
\label{eq:overlap}
\end{equation}
the fraction of claim content terms that also appear in the routed premise. Rescue requires this score to exceed a fixed threshold. This rescue rule is restricted to exact structured evidence; otherwise neutral remains not enough evidence.

\subsection{Calibrated Claim Decision}

The previous stages carry the main verification burden: the router identifies which source is most likely responsible for the claim, NLI determines whether the routed premise entails the claim, and alignment confirms that tokens and protected values are grounded in that source. Together these signals capture the core evidence for source-aware support and already reject clearly unsupported or misattributed claims. Calibration does not replace this reasoning---it sharpens the operating boundary. Because route score, NLI label, and alignment features interact in ways that vary across trace categories and claim types, a fixed decision rule can be overly conservative or permissive in specific regimes. The calibrator learns where that boundary should sit for each combination of signals, producing a more confident and consistent claim decision without changing what the router or NLI model computes.

\begin{figure*}[t]
  \centering
  \begin{tikzpicture}[
  x=1cm,
  y=1cm,
  box/.style={draw, rounded corners=2pt, align=center, minimum width=2.75cm, minimum height=0.72cm, font=\scriptsize},
  data/.style={box, fill=blue!6},
  proc/.style={box, fill=green!7},
  outbox/.style={box, fill=gray!10},
  arrow/.style={-{Latex[length=2.0mm]}, thick}
]
\node[data] (route) at (0,0) {Routing\\score, margin, tool};
\node[data] (nli) at (0,-1.05) {NLI relation\\and score};
\node[data] (align) at (0,-2.10) {Lexical, token,\\protected-value features};
\node[proc] (pre) at (3.25,-1.05) {One-hot\\+ scaling};
\node[proc] (rf) at (6.15,-1.05) {RandomForest\\support calibrator};
\node[outbox] (score) at (9.20,-0.55) {$p_{\mathrm{sup}}(c_j,\hat{s}_j)$};
\node[outbox] (verdict) at (9.20,-1.65) {Support verdict\\at threshold 0.65};

\draw[arrow] (route.east) -- (pre.west);
\draw[arrow] (nli.east) -- (pre.west);
\draw[arrow] (align.east) -- (pre.west);
\draw[arrow] (pre) -- (rf);
\draw[arrow] (rf.east) -- (score.west);
\draw[arrow] (rf.east) -- (verdict.west);
\end{tikzpicture}

\vspace{0.7em}

\begin{tikzpicture}[x=8.6cm,y=3.6cm]
\draw[gray!30, very thin] (0,0) grid[xstep=0.1,ystep=0.2] (1,1);
\draw[->, thick] (0,0) -- (1.04,0) node[right, font=\small] {Support threshold};
\draw[->, thick] (0,0) -- (0,1.08) node[above, font=\small] {Validation reject metric};
\foreach \x/\lab in {0/0,0.2/0.2,0.4/0.4,0.6/0.6,0.8/0.8,1/1.0} {
  \draw (\x,0) -- (\x,-0.015) node[below, font=\scriptsize] {\lab};
}
\foreach \y/\lab in {0/0,0.2/0.2,0.4/0.4,0.6/0.6,0.8/0.8,1/1.0} {
  \draw (0,\y) -- (-0.015,\y) node[left, font=\scriptsize] {\lab};
}
\draw[thick, color=oursblue]
  plot coordinates {
    (0.01,0.000) (0.10,0.577) (0.20,0.731) (0.30,0.726)
    (0.40,0.733) (0.50,0.735) (0.60,0.815) (0.65,0.841)
    (0.70,0.835) (0.80,0.821) (0.90,0.821) (0.99,0.730)
  };
\foreach \x/\y in {0.01/0.000,0.10/0.577,0.20/0.731,0.30/0.726,0.40/0.733,0.50/0.735,0.60/0.815,0.65/0.841,0.70/0.835,0.80/0.821,0.90/0.821,0.99/0.730} {
  \fill[color=oursblue] (\x,\y) circle[radius=0.012];
}
\draw[thick, color=accentgreen]
  plot coordinates {
    (0.01,0.000) (0.10,0.416) (0.20,0.617) (0.30,0.636)
    (0.40,0.649) (0.50,0.695) (0.60,0.903) (0.65,0.981)
    (0.70,1.000) (0.80,1.000) (0.90,1.000) (0.99,1.000)
  };
\foreach \x/\y in {0.01/0.000,0.10/0.416,0.20/0.617,0.30/0.636,0.40/0.649,0.50/0.695,0.60/0.903,0.65/0.981,0.70/1.000,0.80/1.000,0.90/1.000,0.99/1.000} {
  \draw[color=accentgreen, fill=accentgreen] (\x,\y) rectangle ++(0.014,0.014);
}
\draw[dashed, thick, color=accentred] (0.65,0) -- (0.65,1.05);
\node[anchor=south west, font=\scriptsize, fill=white, inner sep=2pt] at (0.03,0.05) {\textcolor{oursblue}{\rule{1.2em}{0.7pt}} Reject F1 \quad \textcolor{accentgreen}{\rule{1.2em}{0.7pt}} Reject recall};
\node[anchor=south, font=\scriptsize, color=accentred] at (0.65,1.05) {0.65};
\end{tikzpicture}
  \caption{Calibration layer. The calibrator receives only verifier-internal routing, NLI, lexical, token-alignment, and protected-value features. It returns \(p_{\mathrm{sup}}\), the probability that claim \(c_j\) is supported by routed source \(\hat{s}_j\). The selected validation threshold, 0.65, converts that probability into a supported-versus-rejected claim decision. The lower panel shows the validation threshold sweep used to select the retained support threshold.}
  \label{fig:calibration}
\end{figure*}
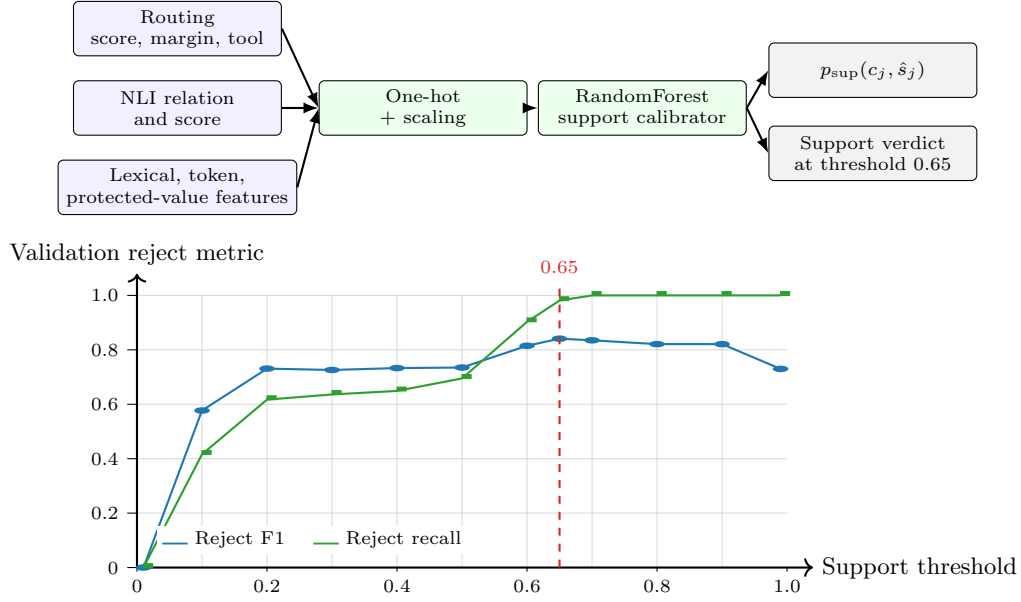

The calibrated claim decision is therefore the support decision for the routed source. Figure~\ref{fig:calibration} shows how the calibrator converts verifier-internal routing, NLI, lexical, alignment, and protected-value signals into an operating threshold. It keeps the source identity fixed and learns an operating boundary from development data using only verifier-internal features. Calibration should tune a verifier whose raw routing and support scores are already meaningful; it should not replace source-aware modeling. The categorical inputs are the NLI label, trace category, predicted tool type, and router status. The numeric inputs are route score, NLI score, routing margin, lexical overlap with the routed source, lexical overlap with all trace evidence, claim length, evidence-chunk count, protected-value counts, missing protected-value counts, stated-attribution indicator, and interaction features such as NLI-score times route-score and route-score times lexical overlap. These inputs are not source-blind baseline outputs; baseline systems are evaluated later and are never used to train or calibrate \ours{}. Appendix~\ref{app:calibration_json} gives a worked example of the calibrator input and output for a representative claim.

The retained calibrator is a random-forest classifier trained on the training partition of the development data. The target is binary support for the routed source: \(z_j=1\) when claim \(c_j\) is adjudicated supported by \(\hat{s}_j\), and \(z_j=0\) for not-enough-evidence, unsupported, contradiction, or failed-source cases. Categorical features are one-hot encoded and numeric features are standardized before fitting. Table~\ref{tab:calibration_training} summarizes the fitting settings and validation-selected operating point. Because a random forest is not trained by epochs, we report its split objective and threshold sweep rather than a neural loss curve.

\begin{table*}[t]
\centering
\tablefontsize
\makebox[\textwidth][c]{%
\begin{tabular}{@{}ll@{}}
\toprule
Item & Setting or result \\
\midrule
Training target & \parbox[t]{0.76\textwidth}{\raggedright Binary claim support for the routed source: supported \(=1\), any blocked label \(=0\).} \\
Rows & \parbox[t]{0.76\textwidth}{\raggedright 1,597 training claims, 367 validation claims, and 361 held-out claims.} \\
Class balance & \parbox[t]{0.76\textwidth}{\raggedright Training: 1,171 supported and 426 blocked; validation: 213 supported and 154 blocked; held-out: 222 supported and 139 blocked.} \\
Preprocessing & \parbox[t]{0.76\textwidth}{\raggedright One-hot encoding for NLI label, category, predicted tool, and router status; standard scaling for numeric verifier features.} \\
Random forest & \parbox[t]{0.76\textwidth}{\raggedright 400 trees, maximum depth 5, minimum leaf size 8, balanced class weights, Gini split criterion, bootstrap bagging, random seed 20260607.} \\
Threshold selection & \parbox[t]{0.76\textwidth}{\raggedright Validation sweep over thresholds 0.01--0.99; selected threshold 0.65 by reject/block F1 with reject accuracy as the tie-breaker.} \\
Validation result & \parbox[t]{0.76\textwidth}{\raggedright Reject F1 0.841, precision 0.737, recall 0.981, accuracy 0.845.} \\
Held-out result & \parbox[t]{0.76\textwidth}{\raggedright Reject F1 0.802, precision 0.673, recall 0.993, accuracy 0.812; reject/block F1 95\% trace-bootstrap CI [0.664, 0.900].} \\
\bottomrule
\end{tabular}
}
\caption{Training details for the retained calibrated claim decision. The random forest is summarized by its fitting settings rather than by an epoch-wise neural loss curve; its split objective is Gini impurity, and the operating point is selected by the validation threshold curve in Figure~\ref{fig:calibration}.}
\label{tab:calibration_training}
\end{table*}

After fitting, the forest estimates
\begin{equation}
p_{\mathrm{sup}}(c_j,\hat{s}_j) = P(z_j=1 \mid \phi(c_j,\hat{s}_j,E)),
\label{eq:calibration}
\end{equation}
where \(\phi(c_j,\hat{s}_j,E)\) is the feature vector assembled from routing, NLI, alignment, lexical, and protected-value checks for claim \(c_j\), routed source \(\hat{s}_j\), and evidence set \(E\). The support threshold is selected on validation by maximizing reject/block F1, with reject accuracy used only as a tie-breaker. The selected threshold is 0.65. A claim is passed forward as supported only when \(p_{\mathrm{sup}}(c_j,\hat{s}_j)\geq 0.65\); otherwise it is passed forward as rejected for the routed source.

This calibration step deliberately answers only one question: \emph{is the claim supported by the routed MCP source?} It does not yet decide whether the answer described that source correctly. The result passed to the next stage is a tuple containing the calibrated support decision, the routed source identifier, and the evidence features that explain the decision.

\subsection{Attribution and Conflation}

Attribution is the second decision. Once calibration has decided whether claim \(c_j\) is supported by routed source \(\hat{s}_j\), \ours{} compares that routed source with the source family stated or implied in the answer. Explicit attribution spans are matched by lexical aliases for record, search, database, document, metadata, and tool-name variants. If no explicit span is present, domain rules can supply defaults for source families that are unambiguous in the trace; otherwise ambiguous claims are marked unavailable rather than silently assigned. A claim can be supported by some MCP source and still be wrong as an attributed answer if the answer assigns it to a different source. Let \(a_j\) be the source family stated or implied by claim \(c_j\), and let \(\hat{s}_j\) be the routed supporting source. \ours{} marks a source conflation when the calibrated support decision is positive but \(a_j\) and \(\hat{s}_j\) refer to incompatible provenance families.

For example, a plan-specific account fact may be supported by an account-record tool output but incorrectly introduced as a policy-document statement. The factual content is then not the only issue: the answer has also misrepresented where the fact came from. Appendix~\ref{app:example_json} gives a complete worked example including routing, NLI scoring, attribution detection, and repair output for a source-conflation case.

\subsection{Answer Decision}

\ours{} aggregates claim decisions with a fail-closed policy. An answer is blocked if any claim has source conflation, high-confidence contradiction, missing protected values, failed routing, or insufficient support. An answer is allowed only when every factual claim is supported by the appropriate routed source. Empty evidence, malformed claims, or failed verifier components are not silently accepted.

\subsection{Repair and Reverification}

Blocked answers enter a bounded revise-and-reverify loop. The repair step receives the original answer, the claim-level verifier outputs, and the routed evidence. It can rewrite unsupported spans, correct source attribution, or remove claims that cannot be grounded. The reported implementation is RARR-style: it follows the retrieve/revise/reverify pattern, but many blocked full-trace cases terminate through deterministic evidence-only rewrites, blocked-claim pruning, or fallback text that makes no evidence-requiring factual claim rather than the original open-web RARR procedure. The revised answer is then evaluated by the same \ours{} verifier.

The loop terminates when the revised answer passes source-aware verification or when remaining unverified content is replaced by a response that contains no factual claim requiring tool evidence. RARR is therefore not treated as a separate oracle; it is a repair mechanism whose output must satisfy the same attribution-sensitive checks that blocked the original answer.

\section{Experimental Setup}
\label{sec:setup}

\subsection{Medical MCP-Agent Use Case}

We evaluate the pipeline in Section~\ref{sec:method} on real traces from a clinical decision support system. The system consists of an orchestrator core connected to four MCP servers: a patient-record tool, a FHIR-like record resource, a literature-search tool (PubMed-style search), and a metadata tool. The medical agent is an instantiation of the trace interface, not part of the definition of \ours{}. The setting is useful for evaluation because the four sources produce adjacent but distinct provenance objects.

The full trace set is used for answer-level and repair evaluation. A trace-level split with \adjlabels{} claims is used for claim-level support and source-attribution scoring.

\paragraph{Data governance.}
The frozen traces are internal MCP-agent evaluation traces collected for this benchmark. Patient-like records, names, identifiers, and appendix examples are synthetic or de-identified benchmark artifacts; no protected health information is included in the released artifact bundle. The study evaluates verifier behavior on captured tool outputs and is not a clinical intervention or a medical-device validation study.

\subsection{Claim Labels}

The claim-level evaluation follows the decomposition step in Section~\ref{sec:method}. Answers are decomposed into atomic claims, and each claim receives a \adjlabels{} support label, relation label, and best-source annotation. Splits are made by trace rather than by claim so that claims from the same assistant answer do not appear in both development and held-out evaluation.

The held-out split contains 40 traces and 361 claims. Source metrics are computed only for claims with an adjudicated source target. Binary reject/block metrics treat unsupported, not-enough-evidence, contradiction, and conflation outcomes as claims that should be blocked from the answer; supported claims are the allow class.

\subsection{Multi-Source Adjudicated Benchmark Extension}

We also rerun \ours{} on a harder multi-source adjudicated benchmark. This packet is harder than the primary 40-trace held-out split because it is represented as pairwise claim--candidate-source rows: each claim case may have multiple candidate sources, including same-topic distractors. The fixed policy combines the benchmark training split with earlier two-judge positive-boost rows, while validation and test use only the benchmark's adjudicated rows. The locked test split contains 59 questions, 254 claim cases, and 2,587 pairwise source-candidate rows.

The benchmark is useful for the multi-tool cases that the primary held-out split underrepresents. It contains chart-plus-literature cases, literature-summary versus exact-citation cases, same-topic wrong-chart candidates, and count/resource-summary claims. We report these as stress slices rather than as separate training objectives.

The router+NLI rerun is scored on frozen independently extracted answer claims associated with the same locked test questions. This creates a small unit mismatch: the pairwise benchmark split has 254 claim cases, while the frozen extraction artifact contains 263 extracted claims for those 59 questions. We therefore report both counts and make clear which unit each table uses.

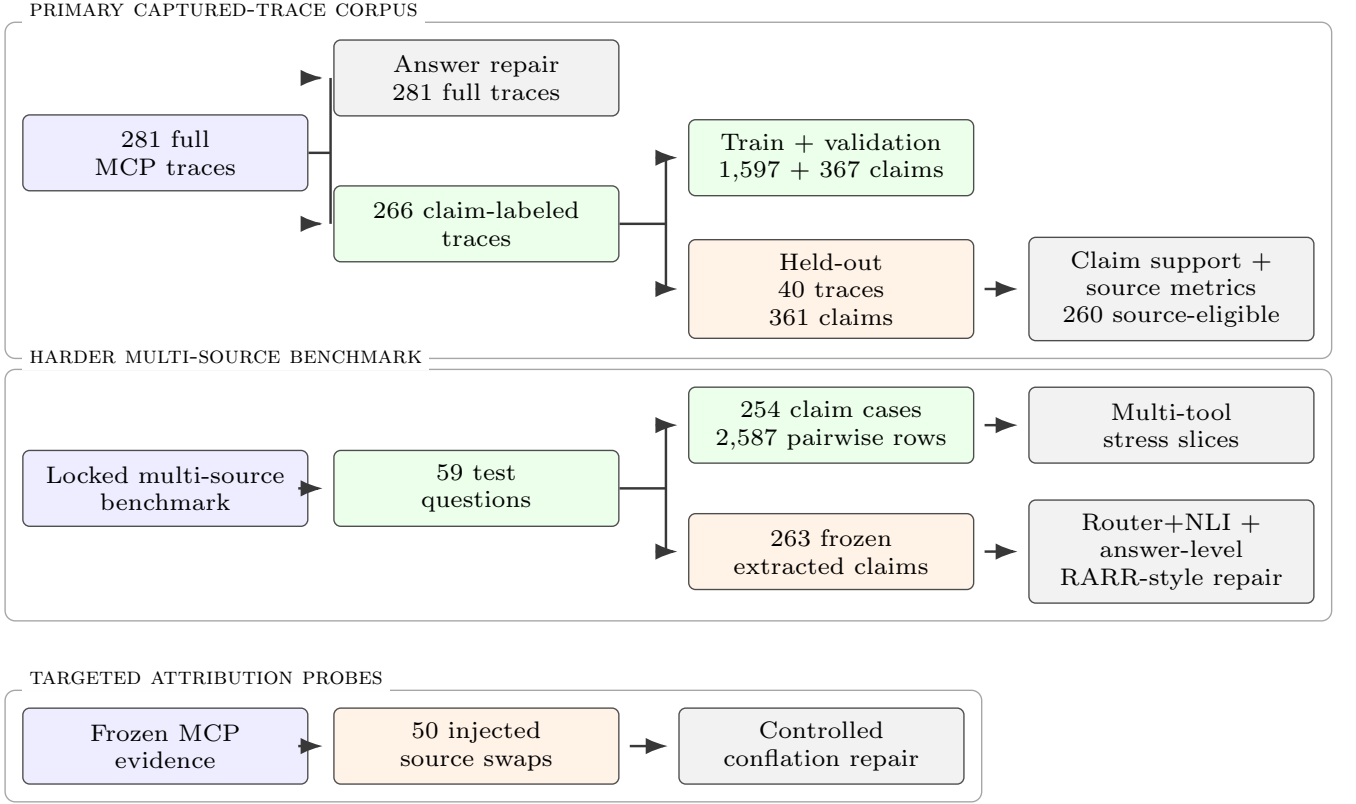
\begin{figure*}[t]
  \centering
  \resizebox{0.98\textwidth}{!}{%
\begin{tikzpicture}[
  x=1cm,
  y=1cm,
  nodebox/.style={draw=black!70, rounded corners=2pt, align=center, text width=2.65cm, minimum height=0.78cm, inner sep=4pt, font=\scriptsize},
  corpus/.style={nodebox, fill=blue!7},
  split/.style={nodebox, fill=green!8},
  eval/.style={nodebox, fill=orange!10},
  output/.style={nodebox, fill=gray!10},
  route/.style={line width=0.72pt, draw=black!78},
  arrow/.style={route, -{Latex[length=2.3mm,width=1.7mm]}, shorten >=3pt, shorten <=3pt},
  brace/.style={draw=black!35, rounded corners=3pt, inner sep=5pt},
  label/.style={font=\scriptsize\scshape, fill=white, inner sep=2pt}
]

\node[corpus] (full) at (0.0,4.15) {281 full\\MCP traces};
\node[output] (rarrfull) at (3.2,4.92) {Answer repair\\281 full traces};
\node[split] (claims) at (3.2,3.42) {266 claim-labeled\\traces};
\node[split] (dev) at (6.85,4.10) {Train + validation\\1,597 + 367 claims};
\node[eval] (heldout) at (6.85,2.75) {Held-out\\40 traces\\361 claims};
\node[output] (claimmetrics) at (10.35,2.75) {Claim support +\\source metrics\\260 source-eligible};

\node[corpus] (locked) at (0.0,0.70) {Locked multi-source\\benchmark};
\node[split] (questions) at (3.2,0.70) {59 test\\questions};
\node[split] (pairwise) at (6.85,1.35) {254 claim cases\\2,587 pairwise rows};
\node[eval] (frozen) at (6.85,0.05) {263 frozen\\extracted claims};
\node[output] (stress) at (10.35,1.35) {Multi-tool\\stress slices};
\node[output] (rerun) at (10.35,0.05) {Router+NLI +\\answer-level RARR-style repair};

\node[corpus] (conflation) at (0.0,-1.95) {Frozen MCP\\evidence};
\node[eval] (swaps) at (3.2,-1.95) {50 injected\\source swaps};
\node[output] (conflationeval) at (6.75,-1.95) {Controlled\\conflation repair};

\coordinate (fullfork) at (1.70,4.15);
\coordinate (fullforktop) at (1.70,4.92);
\coordinate (fullforkbot) at (1.70,3.42);
\draw[route] (full.east) -- (fullfork);
\draw[route] (fullforktop) -- (fullforkbot);
\draw[arrow] (fullforktop) -- (rarrfull.west);
\draw[arrow] (fullforkbot) -- (claims.west);

\coordinate (claimfork) at (5.15,3.42);
\coordinate (claimforktop) at (5.15,4.10);
\coordinate (claimforkbot) at (5.15,2.75);
\draw[route] (claims.east) -- (claimfork);
\draw[route] (claimforktop) -- (claimforkbot);
\draw[arrow] (claimforktop) -- (dev.west);
\draw[arrow] (claimforkbot) -- (heldout.west);
\draw[arrow] (heldout) -- (claimmetrics);

\draw[arrow] (locked) -- (questions);
\coordinate (benchfork) at (5.15,0.70);
\coordinate (benchforktop) at (5.15,1.35);
\coordinate (benchforkbot) at (5.15,0.05);
\draw[route] (questions.east) -- (benchfork);
\draw[route] (benchforktop) -- (benchforkbot);
\draw[arrow] (benchforktop) -- (pairwise.west);
\draw[arrow] (benchforkbot) -- (frozen.west);
\draw[arrow] (pairwise) -- (stress);
\draw[arrow] (frozen) -- (rerun);

\draw[arrow] (conflation) -- (swaps);
\draw[arrow] (swaps) -- (conflationeval);

\node[brace, fit=(full) (claims) (dev) (heldout) (claimmetrics) (rarrfull)] (primarybox) {};
\node[label, anchor=west] at ($(primarybox.north west)+(0.18,0.12)$) {primary captured-trace corpus};
\node[brace, fit=(locked) (questions) (pairwise) (frozen) (stress) (rerun)] (benchbox) {};
\node[label, anchor=west] at ($(benchbox.north west)+(0.18,0.12)$) {harder multi-source benchmark};
\node[brace, fit=(conflation) (swaps) (conflationeval)] (confbox) {};
\node[label, anchor=west] at ($(confbox.north west)+(0.18,0.12)$) {targeted attribution probes};

\end{tikzpicture}%
}
  \caption{Evaluation datasets and units used in the paper. The primary real-trace corpus supports claim-level scoring, source-blind baseline comparison, and full-trace repair. The newer locked multi-source benchmark is reported separately because its pairwise source-candidate rows and frozen extracted claims use different units. A targeted 50-case source-swap probe isolates explicit attribution errors.}
  \label{fig:dataset_map}
\end{figure*}

Figure~\ref{fig:dataset_map} maps the three evaluation objects used below: the primary real-trace corpus, the locked multi-source benchmark, and the targeted source-swap probes. We refer back to these units in the results because claim-level support metrics, pairwise source-candidate metrics, and answer-level repair metrics are not interchangeable.

\subsection{\ours{} Instantiation}
\label{sec:setup-system}

The reported local configuration instantiates Section~\ref{sec:method} with the run constants in Table~\ref{tab:operating_constants}. The table records model identifiers, token budgets, non-learned thresholds, and the retained validation-selected support threshold without repeating the method pipeline.

\begin{table}[!htbp]
\centering
\scriptsize
\setlength{\tabcolsep}{3pt}
\begin{tabular}{@{}ll@{}}
\toprule
Constant & Reported setting \\
\midrule
\parbox[t]{0.28\columnwidth}{\raggedright Maximum extracted claims} & \parbox[t]{0.65\columnwidth}{\raggedright 20 in the service default; artifact-specific scripts report overrides where used.} \\
\parbox[t]{0.28\columnwidth}{\raggedright Source embedding model} & \parbox[t]{0.65\columnwidth}{\raggedright all-MiniLM-L6-v2 SentenceTransformer, cosine similarity over normalized source embeddings.} \\
\parbox[t]{0.28\columnwidth}{\raggedright Chunking} & \parbox[t]{0.65\columnwidth}{\raggedright Complete MCP tool outputs grouped by stable source ID; claim-relevant NLI window capped at 1,500 characters before tokenization.} \\
\parbox[t]{0.28\columnwidth}{\raggedright NLI model} & \parbox[t]{0.65\columnwidth}{\raggedright MoritzLaurer/DeBERTa-v3-base-mnli-fever-anli.} \\
\parbox[t]{0.28\columnwidth}{\raggedright NLI max length} & \parbox[t]{0.65\columnwidth}{\raggedright 512 tokens for new reported Router+NLI reruns; historical 256-token artifacts are labeled as historical.} \\
\parbox[t]{0.28\columnwidth}{\raggedright Alignment threshold \(\tau\)} & \parbox[t]{0.65\columnwidth}{\raggedright 0.35 of the maximum claim-token attention-derived score.} \\
\parbox[t]{0.28\columnwidth}{\raggedright Minimum supported-token ratio} & \parbox[t]{0.65\columnwidth}{\raggedright 0.70 for entailment claims with token-alignment output.} \\
\parbox[t]{0.28\columnwidth}{\raggedright Lexical rescue threshold} & \parbox[t]{0.65\columnwidth}{\raggedright 0.55 content-term overlap when protected values are present; 0.85 otherwise.} \\
\parbox[t]{0.28\columnwidth}{\raggedright Protected-value normalization} & \parbox[t]{0.65\columnwidth}{\raggedright Lowercase alphanumeric, decimal, percentage, date, time, currency, and structured identifier tokens; embedded digits inside alphanumeric words are ignored.} \\
\parbox[t]{0.28\columnwidth}{\raggedright Calibration model and threshold} & \parbox[t]{0.65\columnwidth}{\raggedright RandomForestClassifier, 400 trees, maximum depth 5, minimum leaf size 8, balanced class weights, threshold 0.65 selected on validation by reject/block F1.} \\
\bottomrule
\end{tabular}
\caption{Operating constants for the reported \ours{} configuration. These values make explicit the non-learned thresholds and model identifiers used in the frozen run.}
\label{tab:operating_constants}
\vspace{-0.7\baselineskip}
\end{table}

This setup is designed for offline or data-sensitive review, where the cost of false attribution can dominate latency concerns. It therefore favors conservative blocking and explicit source preservation over a low-latency best-effort answer.

\vspace{-1.4\baselineskip}
\subsection{Claim Decomposition Measurement}

The current artifacts do not contain a separately human-authored gold decomposition set with span-level or proposition-level recall. To make the decomposition stage measurable, we compare the deterministic rule-based decomposer against the frozen independent sentence-level extraction artifact on the locked benchmark test questions. We report this as reference-set agreement, not as human gold decomposition accuracy. The metric is still useful because it exposes over-splitting, missed extracted claims, and protected-value preservation errors before the routing and NLI stages.

\subsection{External Baselines}

MiniCheck, RAGAS Faithfulness, AlignScore, and SummaC-ZS are run on the same held-out claim packet as support comparators. They are compared on binary support metrics only because they do not emit MCP source identifiers.

\vspace{-1.0\baselineskip}
\subsection{Repair and Conflation Evaluation}

RARR-style repair is evaluated as the repair-and-reverification stage described in Section~\ref{sec:method}. Revised answers are scored by the same verifier.

We report a full-trace repair run and a targeted source-conflation slice. The targeted slice injects one deliberate attribution error into otherwise real evidence: patient-record facts are attributed to literature, or literature facts are attributed to the patient chart.

\subsection{Adjudication Status}

The 2,325 claim labels are produced by two independent model judge passes with priority adjudication for disagreements. Human expert review is limited to the final 361-label held-out packet used for the reported primary evaluation.

\section{Results}
\label{sec:results}

\subsection{RQ1: Source-Aware Support Decisions}

\textbf{Finding.} \ours{} is very effective as a fail-closed gate for unsupported claims while also reporting claim-to-source attribution metrics.

\begin{table}[!htbp]
\centering
\scriptsize
\setlength{\tabcolsep}{2pt}
\resizebox{\columnwidth}{!}{%
\begin{tabular}{lrrrrrrrr}
\toprule
Split & Claims & V. acc. & M. F1 & Rej. P & Rej. R & Rej. F1 & Src. & Src+rel \\
\midrule
Train & 1597 & 0.799 & 0.392 & 0.571 & 0.993 & 0.725 & 0.841 & 0.691 \\
Validation & 367 & 0.842 & 0.421 & 0.737 & 0.981 & 0.841 & 0.812 & 0.729 \\
Held-out test & 361 & 0.812 & 0.406 & 0.673 & 0.993 & 0.802 & 0.858 & 0.681 \\
\bottomrule
\end{tabular}
}
\caption{Frozen captured real-agent trace claim-level \ours{} results. Reject precision, recall, and F1 are the operational block metrics: unsupported, contradicted, not-enough-evidence, and conflation are treated as claims to reject, while supported is the allow class. Verdict macro F1 is stricter because it scores the exact factual verdict subtype rather than the binary allow/reject action. Source accuracy and source-plus-relation accuracy are computed over source-eligible claims; the held-out split has 260 source-eligible claims.}
\label{tab:real_trace_router_nli}
\end{table}

\begin{figure}[!htbp]
\centering
\begin{tikzpicture}[font=\small]
  \fill[green!25]  (0,0) rectangle (2.45,1.7);
  \fill[red!20]    (2.45,0) rectangle (4.9,1.7);
  \fill[red!10]    (0,1.7) rectangle (2.45,3.4);
  \fill[green!40]  (2.45,1.7) rectangle (4.9,3.4);
  \draw[thick] (0,0) rectangle (2.45,1.7);
  \draw[thick] (2.45,0) rectangle (4.9,1.7);
  \draw[thick] (0,1.7) rectangle (2.45,3.4);
  \draw[thick] (2.45,1.7) rectangle (4.9,3.4);
  \draw[very thick] (0,0) rectangle (4.9,3.4);
  \node[font=\large\bfseries] at (1.225,1.12) {155};
  \node[font=\small\itshape] at (1.225,0.55) {True Allow};
  \node[font=\large\bfseries] at (3.675,1.12) {67};
  \node[font=\small\itshape] at (3.675,0.55) {False Block};
  \node[font=\large\bfseries] at (1.225,2.82) {1};
  \node[font=\small\itshape] at (1.225,2.25) {False Allow};
  \node[font=\large\bfseries] at (3.675,2.82) {138};
  \node[font=\small\itshape] at (3.675,2.25) {True Block};
  \node[above, font=\scriptsize\bfseries, align=center, text width=2.2cm] at (1.225,3.45) {Predicted\\Allow};
  \node[above, font=\scriptsize\bfseries, align=center, text width=2.2cm] at (3.675,3.45) {Predicted\\Reject};
  \node[anchor=east, font=\scriptsize\bfseries, align=right, text width=1.65cm] at (-0.18,2.55) {Gold\\Reject};
  \node[anchor=east, font=\scriptsize\bfseries, align=right, text width=1.65cm] at (-0.18,0.85) {Gold\\Allow};
  \node[font=\bfseries] at (2.45,4.25) {Predicted Action};
  \node[font=\bfseries, rotate=90] at (-1.65,1.7) {Gold Label};
\end{tikzpicture}
\caption{Held-out binary allow/reject confusion matrix for \ours{} over 361 claims. True Block (138/139) shows near-perfect recall on claims that should be rejected; False Block (67/222) reflects the conservative operating point that sends supported claims to review.}
\label{fig:confusion_matrix}
\end{figure}

Table~\ref{tab:real_trace_router_nli} reports the held-out claim-level metrics, and Figure~\ref{fig:confusion_matrix} visualizes the allow/reject confusion matrix. In these results, ``reject/block'' is the binary decision that a claim should not pass as supported by its routed source. It groups unsupported, contradicted, not-enough-evidence, and conflation cases as the reject class; supported claims are the allow class. On the held-out split, \ours{} reaches 0.802 reject/block F1 with reject recall of 0.993 and reject precision of 0.673. Figure~\ref{fig:confusion_matrix} shows the practical effect: out of 139 claims that should be rejected, the verifier rejects 138 and allows 1; out of 222 supported claims, it allows 155 and sends 67 to review or repair.

\ours{} keeps the routed source ID attached to each claim, giving 0.858 source accuracy over 260 source-eligible held-out claims and 0.681 source-plus-relation accuracy. In deployment terms, the verifier is not only a factuality filter; it also produces a claim-to-source audit trail that can explain why a claim was allowed, blocked, or sent to repair. This supports RQ1: source-aware verification can match or improve support detection while adding a provenance metric.

The retained operating point emphasizes coverage of unsupported content. Its reject precision reflects a review-oriented threshold: some supported claims are held for review, but the policy is intentional for data-sensitive deployment and can be retuned when review burden is the binding constraint. Fine-grained verdict typing is harder than binary rejection: verdict macro F1 is lower because it scores the exact four-way subtype rather than the allow/reject action, so a correctly blocked claim may still receive the wrong reject subtype. The trace-bootstrap interval for reject/block F1 is [0.664, 0.900], reflecting the small 40-trace held-out split and claim clustering within traces.

\FloatBarrier

\subsection{Multi-Source Benchmark Rerun and Multi-Tool Stress Slices}

\textbf{Finding.} On the harder multi-source adjudicated benchmark, \ours{} remains strong at conservative rejection but source-exact attribution is much harder.

\begin{table}[!htbp]
\centering
\scriptsize
\setlength{\tabcolsep}{2pt}
\begin{tabular}{@{}lrrrrrrr@{}}
\toprule
Split & Qs. & Cases & Rows & Claims & Rej. F1 & Src. & Src+rel \\
\midrule
Train & 539 & 3489 & 15248 & -- & -- & -- & -- \\
Validation & 58 & 250 & 2603 & -- & -- & -- & -- \\
Test & 59 & 254 & 2587 & 263 & 0.846 & 0.503 & 0.229 \\
\bottomrule
\end{tabular}
\caption{\ours{} rerun on the locked multi-source adjudicated benchmark. Pairwise cases are the benchmark claim cases; frozen claims are independently extracted answer claims from the same 59 test questions and are the unit scored by the router+NLI rerun.}
\label{tab:v5_provenanceguard_benchmark}
\end{table}

Table~\ref{tab:v5_provenanceguard_benchmark} reports the locked multi-source benchmark rerun. The locked test split contains 254 pairwise claim cases expanded into 2,587 source-candidate rows. The frozen extraction artifact for the same 59 questions contains 263 extracted claims, which are the unit scored by the \ours{} router+NLI rerun. On those frozen extracted claims, \ours{} reaches reject/block F1 0.846, but source accuracy is 0.503 and source-plus-relation accuracy is 0.229. This is the expected direction for a benchmark with many same-topic candidates: binary rejection remains feasible, while exact source ownership becomes substantially harder.

\begin{table}[!htbp]
\centering
\scriptsize
\resizebox{\columnwidth}{!}{%
\begin{tabular}{lr}
\toprule
Slice in locked test benchmark & Cases \\
\midrule
All claim cases & 254 \\
Multiple tool outputs present & 14 \\
Exactly one annotated best source among candidates & 33 \\
Semantically close wrong candidate present & 42 \\
Chart plus literature mixed cases & 64 \\
Literature summary versus exact citation cases & 14 \\
Same-topic but wrong patient/chart source cases & 118 \\
Count or resource-summary claims & 240 \\
\bottomrule
\end{tabular}
}
\caption{Stress slices available in the locked multi-source adjudicated test benchmark, counted after grouping pairwise source-candidate rows by claim case. Slice labels are non-exclusive diagnostics, so counts need not sum to the total number of claim cases.}
\label{tab:v5_multitool_slices}
\end{table}

Table~\ref{tab:v5_multitool_slices} shows that the new benchmark covers failure modes missing from simpler two-source traces. The test split includes 14 multi-tool claim cases, 64 chart-plus-literature mixed cases, 14 literature-summary versus exact-citation cases, 118 same-topic wrong-patient/chart cases, and 240 count or resource-summary cases. It also includes 42 cases with semantically close wrong candidates. These are the cases most relevant to MCP provenance: the wrong source can be topically plausible even when it is not the correct provenance object.

\begin{table}[!htbp]
\centering
\scriptsize
\setlength{\tabcolsep}{2pt}
\resizebox{\columnwidth}{!}{%
\begin{tabular}{lrrrr}
\toprule
Slice & Claims & Rej. F1 & Src. & Src+rel \\
\midrule
Chart + literature mixed & 67 & 0.927 & 0.433 & 0.400 \\
Literature summary vs exact citation & 15 & 0.941 & 0.889 & 0.778 \\
Same-topic wrong chart/patient & 122 & 0.813 & 0.392 & 0.127 \\
Count/resource-summary & 59 & 0.800 & 0.692 & 0.179 \\
\bottomrule
\end{tabular}
}
\caption{\ours{} performance by multi-source stress slice on frozen extracted claims from the locked test questions. Slice rows are small and are intended as diagnostics rather than powered subgroup estimates.}
\label{tab:v5_slice_metrics}
\end{table}

Table~\ref{tab:v5_slice_metrics} shows where exact provenance is hardest. \ours{} retains high reject/block F1 on chart-plus-literature and literature-citation slices, but source-plus-relation accuracy falls to 0.127 on same-topic wrong-chart or wrong-patient cases and 0.179 on count/resource-summary claims. This suggests that source-aware benchmarks should report both support and provenance metrics: a conservative rejector can still fail to identify the exact supporting provenance object.

\begin{table}[!htbp]
\centering
\scriptsize
\setlength{\tabcolsep}{2pt}
\begin{tabular}{@{}lrrrrr@{}}
\toprule
Stage & N & Mean & P50 & P95 & Max \\
\midrule
decide & 263 & \(<0.001\) & \(<0.001\) & \(<0.001\) & \(<0.001\) \\
group\_total & 59 & 0.189 & 0.185 & 0.242 & 0.464 \\
nli & 263 & 0.036 & 0.034 & 0.049 & 0.117 \\
router & 59 & 0.029 & 0.027 & 0.039 & 0.093 \\
\bottomrule
\end{tabular}
\caption{Measured \ours{} claim-packet stage latency on the locked multi-source test questions. The packet provides frozen claims, so decomposition and repair are measured separately in Table~\ref{tab:v5_answer_rarr_latency}.}
\label{tab:v5_pipeline_latency}
\end{table}

Table~\ref{tab:v5_pipeline_latency} reports the claim-packet latency for the same reported local configuration. Mean NLI call latency is 0.036 seconds and mean routing latency per question group is 0.029 seconds, with mean group total 0.189 seconds. The table does not include claim decomposition because the benchmark packet supplies frozen claims, and it does not include repair. Source-aware scoring adds verification time, but for data-sensitive domains the accuracy and attribution check are the priority; end-to-end latency depends primarily on whether the answer enters repair.

\begin{table}[!htbp]
\centering
\scriptsize
\setlength{\tabcolsep}{2pt}
\begin{tabular}{@{}lrrrrrrr@{}}
\toprule
Strategy & Qs. & Pred. & Ref. & Prec. & Rec. & F1 & Value \\
\midrule
Rule-based & 59 & 382 & 263 & 0.644 & 0.935 & 0.763 & 0.563 \\
\bottomrule
\end{tabular}
\caption{Claim-decomposition agreement against the frozen independent sentence-level extraction reference on the locked multi-source test questions. This is reference-set agreement, not a separate human gold span annotation.}
\label{tab:v5_decomposition_accuracy}
\end{table}

Table~\ref{tab:v5_decomposition_accuracy} compares rule-based decomposition with the frozen independent extraction reference. Recall is high (0.935), but precision is lower (0.644), producing 382 claims for 263 reference claims. This is acceptable for a fail-closed verifier because over-splitting tends to increase review burden rather than silently allow unsupported content, but the protected-value exact rate among value-bearing matches is only 0.563. The paper therefore treats decomposition as a measured limitation, not a solved preprocessing step.

\begin{table}[!htbp]
\centering
\scriptsize
\setlength{\tabcolsep}{2pt}
\resizebox{\columnwidth}{!}{%
\begin{tabular}{lrrrrr}
\toprule
Run & Answers & Blocks & Fixed & Allow & Fallback \\
\midrule
Locked multi-source test & 59 & 59 & 59 & 59 & 2 \\
\bottomrule
\end{tabular}
}
\caption{Fresh answer-level \ours{} repair rerun on reconstructed multi-source test traces. The local fail-closed repair policy resolved all initially blocked answers under re-verification; two required the terminal fallback response.}
\label{tab:v5_answer_rarr_summary}
\end{table}

\begin{table}[!htbp]
\centering
\scriptsize
\setlength{\tabcolsep}{2pt}
\begin{tabular}{@{}lrrrrr@{}}
\toprule
Stage & N & Mean & P50 & P95 & Max \\
\midrule
conflation\_check & 902 & \(<0.001\) & \(<0.001\) & \(<0.001\) & \(<0.001\) \\
conflation\_route & 117 & 0.027 & 0.025 & 0.041 & 0.070 \\
decide & 117 & \(<0.001\) & \(<0.001\) & \(<0.001\) & \(<0.001\) \\
decompose & 120 & \(<0.001\) & \(<0.001\) & \(<0.001\) & 0.002 \\
nli & 451 & 0.039 & 0.036 & 0.063 & 0.220 \\
router & 117 & 0.036 & 0.026 & 0.074 & 0.468 \\
\bottomrule
\end{tabular}
\caption{Answer-level RARR rerun latency with rule-based decomposition, local embedding and NLI scoring, one repair iteration, and post-repair re-verification.}
\label{tab:v5_answer_rarr_latency}
\end{table}

Tables~\ref{tab:v5_answer_rarr_summary} and~\ref{tab:v5_answer_rarr_latency} report the fresh answer-level rerun. It reconstructs full test-question traces from the locked benchmark pairwise rows, runs verification with rule-based decomposition, and enables one repair iteration plus re-verification. The external correction editor was unavailable in this local run, so the repair policy used its local fallbacks: 47 evidence-only rewrites, 10 blocked-claim pruning repairs, and 2 terminal fallback responses. All 59 reconstructed answers are initially blocked, all 59 are handled by this repair policy, and all 59 revised outputs pass the same verifier. In this reported local configuration, the rerun averages 0.498 seconds per answer. This is the expected overhead for an offline post-generation gate: for data-sensitive deployments where provenance accuracy matters more than low latency, this cost is appropriate. Production latency will vary with claim count, model placement, and whether repair is invoked.

\subsection{RQ2: Ablation Studies}

\textbf{Finding.} A direct raw verdict head substantially reduces the earlier calibration gap, while routing keeps the correct source near the head of the source ranking.

The initial uncalibrated NLI-only support decision reaches 0.750 reject/block F1 and only 0.363 held-out verdict accuracy, while the retained calibrated \ours{} run reaches 0.802 reject/block F1 and 0.812 verdict accuracy. This 0.449 verdict-accuracy gap showed that the naive raw label mapping was not semantically strong enough. We therefore add a direct raw verdict head trained on the train split over routed Router+NLI and source-evidence features, without selecting a deployment support threshold on validation. This raw verdict head reaches 0.839 held-out verdict accuracy, 0.816 reject/block F1, and 0.704 source-plus-relation accuracy. Threshold calibration on top of this raw head does not improve held-out accuracy: the validation-F1 threshold gives 0.817 verdict accuracy and 0.800 reject/block F1, while a high-recall threshold gives 0.795 verdict accuracy and 0.789 reject/block F1 but restores reject recall to 0.993. The gap to the retained calibrated verifier is therefore no longer a raw-accuracy deficit; it is an operating-point tradeoff. Table~\ref{tab:full_trace_rarr_ablation} also shows that router-only source ranking gives 0.858 Top-1 source accuracy. We do not treat Top-3 or Top-5 as meaningful performance metrics on this split because each source-eligible held-out claim has at most two candidate sources.

\begin{table}[!htbp]
\centering
\scriptsize
\setlength{\tabcolsep}{3pt}
\resizebox{\columnwidth}{!}{%
\begin{tabular}{llr}
\toprule
Condition & Measured outcome & Value \\
\midrule
Router-only source ranking & Top-1 source accuracy over source-eligible claims & 0.858 \\
NLI-only support decision & Reject F1 / reject recall & 0.750 / 0.993 \\
Router+NLI, calibrated & Reject F1 / reject recall & 0.802 / 0.993 \\
Router+NLI, calibrated & Top-1 source accuracy over source-eligible claims & 0.858 \\
\bottomrule
\end{tabular}
}
\caption{Router ablation on the held-out claim split. Router-only reports whether the adjudicated source is ranked first over the 260 source-eligible claims. NLI-only uses the routed NLI support decision without the calibrated evidence-feature layer. Router+NLI, calibrated is the retained \ours{} verifier. Top-3 and Top-5 are omitted because the held-out source candidate sets are too small to make those cutoffs discriminative.}
\label{tab:full_trace_rarr_ablation}
\end{table}

The mechanism is that the naive raw NLI label loses information that is present in evidence-derived features: routing score, routing margin, lexical overlap, token support, and protected-value coverage. The direct raw verdict head learns a verdict decision from those features before any threshold calibration layer is applied. Calibration is still useful for choosing a fail-closed operating point, but it no longer needs to rescue a fundamentally weak raw verdict mapping in this experiment. The calibrated support verdict also creates the precondition for conflation detection: only after the verifier knows which routed source supports the claim can it ask whether the answer attributed that claim to the same provenance family. Hard route-score cutoffs do not solve the problem: Table~\ref{tab:route_threshold_diagnostic} shows that increasingly strict cutoffs reject most claims and reduce source accuracy among retained source-eligible claims.

This confirms the RQ2 hypothesis in a bounded way. Routing alone is necessary for attribution but does not decide support; NLI alone estimates support but does not preserve the source-ranking behavior needed for attribution; calibrated Router+NLI combines both. The limitation is that this held-out split cannot evaluate large-\(k\) source recall: Top-3 and Top-5 would be tautological because each claim has at most two candidate sources. A larger trace set with more simultaneous MCP sources is needed to measure whether top-\(k\) routing remains strong under heavier source competition.

Table~\ref{tab:development_verifier_runs} compares the retained verifier with development variants that test alternative calibration and routing choices. The raw verdict head has the strongest held-out verdict accuracy and reject/block F1, while \ours{} is retained as the fail-closed deployment point because it preserves the highest reject recall among the strong variants. Thresholding the raw head moves it toward that conservative operating point but reduces held-out accuracy and reject/block F1. The evidence-calibrated ExtraTrees/logistic blend and calibrated-score ensemble improve some validation statistics, but in the held-out audit they reduce reject recall and do not improve reject/block F1 relative to the retained system. We therefore report both the raw verdict head as the strongest uncalibrated verifier and \ours{} as the retained conservative verifier for RARR-style repair.

\begin{table}[!htbp]
\centering
\scriptsize
\setlength{\tabcolsep}{2pt}
\resizebox{\columnwidth}{!}{%
\begin{tabular}{lrrrrr}
\toprule
Variant & V. acc. & Rej. R & Rej. F1 & Src. & Src+rel \\
\midrule
Uncalibrated Router+NLI & 0.363 & 0.993 & 0.750 & 0.858 & 0.615 \\
Route-threshold rerun & 0.346 & 0.993 & 0.738 & 0.858 & 0.592 \\
No-source top-$k$ & 0.729 & 0.993 & 0.738 & 0.858 & 0.481 \\
Claim-relevant windows & 0.421 & 0.942 & 0.771 & 0.862 & 0.673 \\
Raw verdict head & 0.839 & 0.928 & 0.816 & 0.858 & 0.704 \\
Raw head + val threshold & 0.817 & 0.950 & 0.800 & 0.858 & 0.688 \\
Raw head + high-recall threshold & 0.795 & 0.993 & 0.789 & 0.858 & 0.685 \\
Evidence-calibrated blend & 0.817 & 0.957 & 0.801 & 0.858 & 0.685 \\
Calibrated-score ensemble & 0.809 & 0.957 & 0.794 & 0.858 & 0.681 \\
\textbf{ProvenanceGuard (ours)} & \textbf{0.812} & \textbf{0.993} & \textbf{0.802} & \textbf{0.858} & \textbf{0.681} \\
\bottomrule
\end{tabular}
}
\caption{Held-out results for source-aware verifier development runs. Reject metrics are the binary block metrics used for support gating. All rows are evaluated on the same 361-claim held-out packet. Source metrics are computed on the 260 tool-source-eligible claims. The retained system is bolded.}
\label{tab:development_verifier_runs}
\end{table}

\begin{table}[!htbp]
\centering
\scriptsize
\setlength{\tabcolsep}{2pt}
\resizebox{\columnwidth}{!}{%
\begin{tabular}{llrrrrr}
\toprule
System & Stage & Rel. & Src. & Src+rel & Rej. F1 & V. acc. \\
\midrule
\textbf{PG legacy pairwise} & \textbf{diagnostic} & \textbf{0.496} & \textbf{0.472} & \textbf{0.212} & \textbf{0.869} & \textbf{0.249} \\
Long DeBERTa & raw & 0.750 & 0.707 & 0.478 & 0.868 & 0.665 \\
Long DeBERTa & calibrated & 0.750 & 0.688 & 0.478 & 0.874 & 0.657 \\
ModernBERT & raw & 0.771 & 0.669 & 0.478 & 0.873 & 0.650 \\
ModernBERT & calibrated & 0.750 & 0.656 & 0.459 & 0.875 & 0.661 \\
ModernCE & raw & 0.681 & 0.675 & 0.420 & 0.814 & 0.626 \\
ModernCE & calibrated & 0.670 & 0.586 & 0.382 & 0.850 & 0.598 \\
\bottomrule
\end{tabular}
}
\caption{Raw and validation-calibrated systems from the current post-finetune benchmark package on the locked multi-source adjudicated benchmark. This table uses the package's pairwise comparison units. The legacy pairwise-scorer row is a diagnostic from the package and is not the retained frozen-claim \ours{} pipeline; Table~\ref{tab:v5_provenanceguard_benchmark} separately reports that frozen extracted-claim rerun.}
\label{tab:v5_raw_calibrated_comparison}
\end{table}

Table~\ref{tab:v5_raw_calibrated_comparison} clarifies the role of calibration on the multi-source benchmark. Calibration is useful when it tunes an already meaningful score into an operating threshold. It is not a substitute for a raw model that understands source ownership. In the current post-finetune benchmark package, the completed two-head backbones already have much stronger raw source-plus-relation accuracy than the legacy comparison rows: Long DeBERTa and ModernBERT reach 0.478, and ModernCE reaches 0.420. Validation calibration changes the operating point rather than uniformly improving every metric: reject/block F1 increases slightly for the calibrated variants, while source-plus-relation accuracy is unchanged or lower. This supports the design goal for future versions: improve the base verifier so raw relation and source scores are sensible, then use calibration only to choose a deployment operating point.

\begin{table}[!htbp]
\centering
\scriptsize
\setlength{\tabcolsep}{2pt}
\resizebox{\columnwidth}{!}{%
\begin{tabular}{lrrrrl}
\toprule
Model & Tokens & Rows & Status & Src+rel & Note \\
\midrule
ModernBERT-base raw & 2048 & 2587 & complete & 0.090 & 100-group 1-epoch \\
\bottomrule
\end{tabular}
}
\caption{Raw longer-context diagnostic using a 2048-token ModernBERT checkpoint trained for one bounded 100-group epoch.}
\label{tab:v5_long_context_raw}
\end{table}

Table~\ref{tab:v5_long_context_raw} reports the replacement long-context diagnostic. After the original 20-epoch 2048-token ModernBERT checkpoint proved unreadable, we trained a replacement 2048-token ModernBERT checkpoint for one bounded 100-group epoch and evaluated it raw on the full locked multi-source test split. The result is intentionally a smoke-quality long-context checkpoint rather than a full finetune: case source-plus-relation accuracy is 0.090, with source-pair accuracy 0.559. This gives a valid long-context raw test point, but it does not yet support the stronger claim that longer context alone makes the raw verifier sensible.

Table~\ref{tab:route_threshold_diagnostic} explains why a hard no-source cutoff was not retained. Increasing the cutoff rejects most claims before NLI and reduces source accuracy among the retained source-eligible claims, so route score alone is too blunt for support decisions.

\begin{table}[!htbp]
\centering
\scriptsize
\begin{tabular}{rrrr}
\toprule
Cutoff & Kept & Rejected & Source acc. \\
\midrule
0.100 & 314 & 47 & 0.847 \\
0.200 & 191 & 170 & 0.787 \\
0.300 & 89 & 272 & 0.684 \\
0.400 & 39 & 322 & 0.533 \\
0.500 & 9 & 352 & 0.400 \\
0.600 & 2 & 359 & 0.000 \\
0.700 & 1 & 360 & 0.000 \\
0.800 & 0 & 361 & -- \\
\bottomrule
\end{tabular}
\caption{Held-out route-score threshold diagnostic. A hard no-source cutoff rejects low-score claims before NLI. Higher cutoffs reject most claims and reduce source accuracy among retained source-eligible claims, so this policy was not retained. Source accuracy is undefined when no claims are retained.}
\label{tab:route_threshold_diagnostic}
\end{table}

\begin{table}[!htbp]
\centering
\scriptsize
\setlength{\tabcolsep}{2pt}
\resizebox{\columnwidth}{!}{%
\begin{tabular}{lrrrrrr}
\toprule
Split & Blocked & NLI/NEE & Wrong route & Contradiction & No source & Other \\
\midrule
Train & 741 & 592 & 149 & 0 & 0 & 0 \\
Validation & 205 & 167 & 38 & 0 & 0 & 0 \\
Held-out test & 205 & 176 & 29 & 0 & 0 & 0 \\
\bottomrule
\end{tabular}
}
\caption{Blocked prediction taxonomy for \ours{} on frozen captured real-agent traces.}
\label{tab:real_trace_blocked_taxonomy}
\end{table}

Table~\ref{tab:real_trace_blocked_taxonomy} explains where the system spends its recall. On the held-out split, 176 blocked claims are NLI-neutral or not-enough-evidence cases and 29 are wrong-route candidates. This is consistent with the method design: \ours{} is primarily rejecting claims that cannot be grounded in the routed source, not only overt contradictions.

\FloatBarrier

\subsection{RQ3: Comparison With Source-Blind Baselines}

\textbf{Finding.} Source-blind support baselines are competitive on binary support, but they cannot evaluate MCP source attribution.

Table~\ref{tab:real_trace_external_claim_baselines} reports the source-blind baseline comparison and is read only on the support axis. \ours{} has the highest reject/block F1 at 0.802, followed by MiniCheck at 0.783 and RAGAS Faithfulness at 0.758. MiniCheck is close enough that the 0.019 absolute F1 gap is not statistically significant under paired trace-level bootstrap comparison (one-sided \(p\approx0.13\); two-sided \(p\approx0.26\)). The practical difference is metric coverage: only \ours{} reports claim-to-source IDs, source accuracy, and source-plus-relation accuracy. The table's lower verdict macro F1 values should not be read as a contradiction of high reject/block scores: macro F1 asks for the exact four-way verdict subtype, while reject/block metrics ask only whether the claim should be allowed or rejected.

\begin{table}[!htbp]
\centering
\scriptsize
\setlength{\tabcolsep}{2pt}
\resizebox{\columnwidth}{!}{%
\begin{tabular}{lrrrrrr}
\toprule
System & Claims & Rej. P & Rej. R & Rej. F1 & V. acc. & M. F1 \\
\midrule
\textbf{ProvenanceGuard (ours)} & \textbf{361} & \textbf{0.673} & \textbf{0.993} & \textbf{0.802} & \textbf{0.812} & \textbf{0.406} \\
MiniCheck & 361 & 0.655 & 0.971 & 0.783 & 0.792 & 0.396 \\
RAGAS Faithfulness & 361 & 0.660 & 0.892 & 0.758 & 0.781 & 0.390 \\
AlignScore & 361 & 0.510 & 0.942 & 0.662 & 0.629 & 0.313 \\
SummaC-ZS & 361 & 0.433 & 0.439 & 0.436 & 0.562 & 0.270 \\
\bottomrule
\end{tabular}
}
\caption{Held-out real-agent trace claim-level factuality comparison. Reject metrics are the binary block metrics: they ask whether a claim should be rejected rather than allowed as supported. Verdict macro F1 is macro F1 over the four factual verdict labels: supported, unsupported, contradicted, and not-enough-evidence. \ours{} uses only router, NLI, and evidence-derived features. MiniCheck, RAGAS Faithfulness, AlignScore, and SummaC-ZS are source-blind claim/evidence support baselines and therefore are not scored on claim-to-source attribution.}
\label{tab:real_trace_external_claim_baselines}
\end{table}

Table~\ref{tab:real_trace_uncertainty} confirms that the support-only advantage should not be overread. \ours{} reaches 0.802 reject/block F1 with a 95\% CI of [0.664, 0.900], MiniCheck reaches 0.783 [0.645, 0.882], and RAGAS Faithfulness reaches 0.758 [0.618, 0.861]. These intervals overlap because the held-out split contains only 40 traces.

\begin{table}[!htbp]
\centering
\tablefontsize
\resizebox{\columnwidth}{!}{%
\begin{tabular}{lrr}
\toprule
Quantity & Point estimate & 95\% CI \\
\midrule
\textbf{ProvenanceGuard (ours) reject/block F1} & \textbf{0.802} & \textbf{[0.664, 0.900]} \\
MiniCheck reject/block F1 & 0.783 & [0.645, 0.882] \\
RAGAS Faithfulness reject/block F1 & 0.758 & [0.618, 0.861] \\
AlignScore reject/block F1 & 0.662 & [0.498, 0.778] \\
SummaC-ZS reject/block F1 & 0.436 & [0.287, 0.555] \\
\textbf{ProvenanceGuard (ours) source accuracy} & \textbf{0.858} & \textbf{[0.731, 0.963]} \\
Conflation-probe repair resolution rate & 1.000 & [0.93, 1.00] \\
Post-repair pass among resolved probes & 1.000 & [0.93, 1.00] \\
\bottomrule
\end{tabular}
}
\caption{Uncertainty estimates for held-out claim-level metrics and targeted repair. Claim-level intervals use 5,000 trace-level bootstrap resamples over the 40 held-out traces. Repair intervals for 50/50 probe successes use an exact binomial 95\% interval rounded to two decimals because case-level bootstrap resampling is degenerate on the fixed controlled probe set. \ours{} source accuracy is computed over 260 source-eligible held-out claims.}
\label{tab:real_trace_uncertainty}
\end{table}

The limitation is scope. The baselines are not failed attribution systems; they are different tools. They remain appropriate comparators for support estimation, and their strong held-out F1 shows that support checking is a meaningful baseline. The unexpected result is how close MiniCheck is on support F1, which suggests that the clearest contribution of \ours{} is the ability to retain source ownership while maintaining competitive support detection.

\FloatBarrier

\subsection{RQ4: Repair After Source-Aware Blocking}

\textbf{Finding.} The repair loop can turn blocked real-trace answers into verifier-passing answers, but many repairs require fallback text rather than a fully rewritten substantive answer.

Table~\ref{tab:real_trace_rarr} reports full real-trace repair. On 281 full real traces, the pre-repair verifier allows 108 answers and blocks 173. The repair loop resolves all 173 blocked answers, and all 173 revised outputs pass the same Router+NLI verifier. However, 144 of those resolutions use terminal fallback text. This full-trace result should be separated from the targeted conflation slice in RQ5, where the controlled probes are deliberately simpler.

\begin{table}[!htbp]
\centering
\tablefontsize
\resizebox{\columnwidth}{!}{%
\begin{tabular}{lrrrrr}
\toprule
Rows & Pre allow & Pre block & Repair resolved & Terminal fallback & Revised allow \\
\midrule
281 & 108 & 173 & 173 & 144 & 173 \\
\bottomrule
\end{tabular}
}
\caption{Full real-trace RARR-style repair results on 281 quality-filtered MCP-agent traces. Rejected answers enter the repair loop and are then re-scored by the same source-router plus NLI verifier. Terminal fallback means the remaining unverified content was replaced with text that makes no factual claim requiring tool evidence.}
\label{tab:real_trace_rarr}
\end{table}

The answer-level rerun in Table~\ref{tab:v5_answer_rarr_summary} is a second repair check on the multi-source benchmark rather than on the original 281-trace corpus. It is stricter in the sense that all reconstructed benchmark answers are initially blocked, but it is also smaller and uses rule-based decomposition rather than the LLM decomposition configuration used for the main full-trace run.

The mechanism is fail-closed repair. RARR first attempts source-grounded rewriting, then reruns the verifier. If unsupported content remains, the terminal fallback removes the remaining factual claim rather than allowing an unverifiable answer. This supports the repair hypothesis only in the conservative sense: the loop can prevent unverifiable content from passing, but it does not always recover a rich answer.

The limitation is that repair success is measured against the same verifier that triggered the block. This is appropriate for checking pipeline consistency, but it is not independent proof that the revised answer is complete for the application domain. Compared with prior RARR work, this setting is stricter because revision must preserve MCP source attribution, not only improve general factual consistency against retrieved text.

\FloatBarrier

\subsection{RQ5: Targeted Source Conflation}

\textbf{Finding.} \ours{} detects and repairs deliberately injected source-conflation errors in a controlled provenance stress test.

Table~\ref{tab:doctor_conflation_rarr} reports the targeted source-conflation probes. In 50 probes, the verifier blocks all 50 source-confused replies and labels all 50 as explicit conflations. The repair policy resolves all 50 cases; all 50 pass post-repair verification, and no revised answer retains the deliberately wrong attribution. The exact binomial 95\% interval for 50/50 successes is approximately [0.93, 1.00]; the bootstrap interval is degenerate on this fixed controlled probe set and should not be read as a generalization interval.

\begin{table}[!htbp]
\centering
\tablefontsize
\resizebox{\columnwidth}{!}{%
\begin{tabular}{lrrrrrr}
\toprule
Rows & Pre block & Explicit confl. & Repair resolved & Terminal fallback & Post allow & Wrong attr. retained \\
\midrule
50 & 50 & 50 & 50 & 11 & 50 & 0 \\
\bottomrule
\end{tabular}
}
\caption{Targeted source-conflation repair benchmark over frozen real MCP evidence from the medical use case. The 50 probes contain one deliberate source-attribution error each; post-repair verification uses the same source-router plus NLI verifier.}
\label{tab:doctor_conflation_rarr}
\end{table}

The mechanism is the separation between support and attribution. The factual content in these probes is supported by the trace, but assigned to the wrong source family. \ours{} detects this by comparing stated attribution with the routed supporting source.

This supports the source-conflation hypothesis for explicit attribution swaps. The limitation is difficulty: each probe contains one clear source swap and no adversarial attempt to hide the attribution error. The 1.000 result should therefore be interpreted as a diagnostic check for a constrained failure mode, not as evidence that all source-confusion cases are solved.

\FloatBarrier

\subsection{Adjudication Status}

\textbf{Finding.} The held-out evaluation uses complete \adjlabels{} labels with human expert review.

Table~\ref{tab:model_expert_simulation_status} summarizes the adjudication status. The held-out packet has labels for all 361 claims. The agreement rows reflect two independent judge prompts run on a local Gemma 4 E4B instruction-tuned model and priority adjudication for disagreements. Human experts then reviewed the resulting held-out labels before they were used for evaluation.

\begin{table}[!htbp]
\centering
\tablefontsize
\begin{tabular}{l r}
\toprule
Item & Count \\
\midrule
Held-out claims in adjudication packet & 361 \\
LLM-assisted labeled claims & 361 \\
Model-assisted agreement rows & 350 \\
Priority-adjudicated disagreements & 11 \\
Human expert reviewed held-out labels & 361 \\
\bottomrule
\end{tabular}
\caption{Held-out adjudication status. Labels come from two independent judge prompts run on a local Gemma 4 E4B instruction-tuned model, priority adjudication for disagreements, and human expert review of the resulting 361 held-out labels. This table does not claim human review of the training or validation labels.}
\label{tab:model_expert_simulation_status}
\end{table}

The mechanism is a two-stage adjudication workflow: model judges provide first-pass labels, disagreements receive priority adjudication, and human experts verify the final held-out labels. This supports reproducible benchmarking while incorporating expert review after model adjudication.

\FloatBarrier

\section{Discussion}

The main lesson is not that attribution checks establish domain truth. They do not; they add a provenance-preserving layer on top of support estimation. In MCP settings, this lets the verifier distinguish claims that are merely supported somewhere in the trace from claims that are supported by the source the answer names or implies.

For an agent developer, the practical effect is a post-generation gate. Unsupported or incorrectly attributed claims are blocked before release, and the answer is either repaired, shortened, or replaced by fallback text when support cannot be established. This behavior is intentionally conservative. It is most appropriate for offline or data-sensitive workflows where preventing unsupported or misattributed content is more important than minimizing every increment of verification latency.

The multi-source benchmark sharpens the evaluation target. Binary support detection can remain strong even when exact source ownership is difficult, especially when several candidate sources are semantically close. Provenance-aware evaluation should therefore report both axes: whether a claim should pass at all, and whether the verifier identified the right source family.

The calibration results lead to a practical engineering conclusion. Calibration is useful for selecting an operating threshold, but it should tune a verifier whose raw source and relation signals are already meaningful. Future versions should prioritize stronger source-aware raw models, harder negative candidates, and better decomposition before relying on additional calibration layers. The replacement 2048-token ModernBERT smoke run validates the evaluation path, but it does not yet show that longer context alone solves source ownership.

The repair results should be interpreted as fail-closed pipeline evidence. RARR-style repair plus reverification can remove or correct unsupported content, and the targeted conflation probes show that explicit attribution swaps are detectable. However, repair success against the same verifier is not independent domain validation, and terminal fallback text means the system avoided an unverifiable answer rather than necessarily recovering a rich one.

The reported system was tested with local LLM components, including a Gemma 4 E4B instruction-tuned model for claim decomposition and LLM-assisted adjudication. This keeps the verifier compatible with offline and data-sensitive environments, but it also means that stronger state-of-the-art LLMs remain an untested path for improving decomposition, adjudication consistency, and repair rewriting. We treat that as a generalization hypothesis, not as a reported result.

The main open question is scale. The primary held-out split is small, confidence intervals are wide, and it cannot meaningfully evaluate Top-3 or Top-5 source routing because the current traces usually expose only one or two candidate sources per claim. The multi-source benchmark is a step toward this harder setting, but it also shows that richer source competition requires stronger source-aware models and larger evaluation packets.

\section{Conclusion}

We presented \ours{}, a source-aware factuality verifier for MCP-based LLM agents. The paper's central claim is that factuality verification in multi-tool settings must determine both whether a claim is supported and whether it is attributed to the correct source. This distinction matters because a claim can be supported somewhere in the available MCP trace while still being misleadingly assigned to the wrong tool output, structured record, search result, database entry, or metadata source.

\ours{} preserves stable MCP source identifiers, decomposes answers into atomic claims, routes claims to source-specific evidence, checks support with NLI and alignment signals, and compares stated attribution with the routed source. In the medical MCP testbed used here, the system achieved competitive blocking performance while also producing claim-to-source attribution judgments unavailable from source-blind baselines. In targeted source-conflation probes, it detected all deliberately injected attribution swaps, demonstrating the value of source-aware verification.

These results suggest that source attribution should be treated as an independent evaluation axis for tool-using LLM agents. However, the current evidence is limited by the use of one medical agent stack, labels that are \adjlabels{}, a small held-out split, and controlled conflation probes. Future work should expand evaluation to larger, more diverse MCP environments and include more subtle source-confusion cases.

\section{Limitations}

The trace benchmark is drawn from one medical MCP-agent stack. It is representative of that stack, but it does not establish universal medical factuality or domain safety validation.

The reproducible object is the frozen trace, not future behavior of PubMed, FHIR resources, search tools, or the live agent. Tool outputs and agent behavior may vary if the same prompts are rerun at a later date.

The 2,325-label claim subset uses LLM-assisted adjudication with two independent judge prompts and priority adjudication for disagreements. Human expert verification covers only the 361 held-out labels used in the reported primary evaluation. The benchmark should therefore not be interpreted as a fully clinician-adjudicated training corpus.

Human expert verification covers the held-out labels used in the reported evaluation, but it does not establish universal medical factuality or domain safety validation.

The claim-level bootstrap intervals are wide because the held-out split contains only 40 traces and claims are clustered within traces. Reported uncertainty therefore uses trace-level resampling rather than treating all 361 claims as independent.

The held-out labels are dominated by supported and not-enough-evidence claims. There are few explicit contradictions and no gold conflation relation in the random held-out claim split. The targeted 50-case source-confusion benchmark is therefore the most direct evidence for controlled conflation detection and repair, while the random 281-trace run evaluates full-answer behavior. Because the 50 probes are generated by injecting one explicit source-attribution swap into otherwise real evidence, the 1.000 repair metrics do not imply that the task is solved under harder, multi-error, paraphrased, or adversarial source-confusion conditions.

The paper reports claim-decomposition agreement against a frozen independent extraction artifact, not against a separately human-authored gold atomic-claim set. This makes the decomposition numbers useful for regression testing and error analysis, but not a final estimate of human gold extraction precision and recall. Decomposition errors can still affect answer-level behavior, especially protected-value preservation, so this remains an evaluation gap.

The initial raw Router+NLI label mapping depended heavily on calibration for final verdict accuracy. On the main held-out split, the naive raw verdict accuracy is 0.363 and the retained calibrated verifier reaches 0.812. The direct raw verdict head reduces this gap by reaching 0.839 verdict accuracy without validation-threshold calibration, but it is still trained on split-specific Router+NLI and source-evidence features rather than being an end-to-end source-aware NLI model. Threshold calibration on top of the raw head shows the expected recall--accuracy tradeoff rather than an additional held-out accuracy gain: the high-recall operating point reaches 0.993 reject recall but falls to 0.795 verdict accuracy. This improves the reported calibration-dependence story, while leaving robustness under distribution shift as an evaluation gap.

The 2048-token raw ModernBERT result uses a replacement one-epoch checkpoint trained for only 100 sampled groups after the original longer run produced a corrupt checkpoint archive. This is enough to verify the evaluation path and report a valid raw long-context test point, but it is not a full training run. We therefore do not claim that longer context improved raw source-aware verification.

External source-blind support baselines are not source-attribution systems. They are appropriate binary decision comparators where configured, but they are not baselines for claim-to-source accuracy.

The LLM-dependent components are also a validity boundary. The reported configuration uses a local Gemma 4 E4B instruction-tuned model where LLM calls are required. Stronger frontier LLMs may produce cleaner decompositions, more stable adjudications, or richer repairs, but those systems are not evaluated in the reported frozen run.

\section{Reproducibility}

\begin{center}
\scriptsize
\setlength{\tabcolsep}{3pt}
\refstepcounter{table}\label{tab:reproducibility}
\begin{tabular}{@{}ll@{}}
\toprule
\parbox[t]{0.30\columnwidth}{\raggedright Component} & \parbox[t]{0.64\columnwidth}{\raggedright Reproducibility record} \\
\midrule
\parbox[t]{0.30\columnwidth}{\raggedright Evaluation unit} & \parbox[t]{0.64\columnwidth}{\raggedright Frozen MCP trace set plus the derived claim-level evaluation packet.} \\
\parbox[t]{0.30\columnwidth}{\raggedright Generated outputs} & \parbox[t]{0.64\columnwidth}{\raggedright Reported tables are generated from stored traces, claim packets, \adjlabels{} labels, prediction files, baseline outputs, uncertainty resamples, and repair outputs.} \\
\parbox[t]{0.30\columnwidth}{\raggedright Routing model} & \parbox[t]{0.64\columnwidth}{\raggedright all-MiniLM-L6-v2 SentenceTransformer embeddings with cosine similarity over normalized embeddings.} \\
\parbox[t]{0.30\columnwidth}{\raggedright NLI verifier} & \parbox[t]{0.64\columnwidth}{\raggedright MoritzLaurer DeBERTa-v3-base-mnli-fever-anli with a 512-token budget by default; 256-token artifacts are labeled historical.} \\
\parbox[t]{0.30\columnwidth}{\raggedright Claim decomposition and adjudication} & \parbox[t]{0.64\columnwidth}{\raggedright Local Gemma 4 E4B instruction-tuned model for the reported decomposition and LLM-assisted adjudication configuration.} \\
\parbox[t]{0.30\columnwidth}{\raggedright Calibration} & \parbox[t]{0.64\columnwidth}{\raggedright RandomForestClassifier trained on the development split; the support threshold is selected on validation before held-out scoring.} \\
\parbox[t]{0.30\columnwidth}{\raggedright Repair outputs} & \parbox[t]{0.64\columnwidth}{\raggedright Full-trace repair outputs are stored separately from claim-level prediction outputs.} \\
\parbox[t]{0.30\columnwidth}{\raggedright External baselines} & \parbox[t]{0.64\columnwidth}{\raggedright Baseline outputs are stored as support-only predictions and are not used to train or calibrate \ours{}.} \\
\bottomrule
\end{tabular}
\par\vspace{0.35\baselineskip}
\raggedright TABLE~\thetable. Reproducibility record for the reported evaluation.
\end{center}

Table~\ref{tab:reproducibility} summarizes the frozen artifacts, model settings, and stored outputs used to reproduce the reported evaluation.

\section*{Funding}
\textbf{CHAMELEON file number:} MIE-010100-2024-53

Funding was awarded under the support programme for projects promoting the
microelectronics and semiconductor value chain (PERTE CHIP), as part of
Spain's Recovery, Transformation and Resilience Plan.

\makeatletter
\def\bibsection{%
  \par
  \onecolumngrid@push
  \vspace{0.85\baselineskip}%
  \begingroup
    \centering
    \phantomsection
    \addcontentsline{toc}{section}{References}%
    {\bfseries REFERENCES\par}%
  \endgroup
  \vspace{0.5\baselineskip}%
  \onecolumngrid@pop
}
\makeatother

\bibliographystyle{apsrev4-2}
\bibliography{references}

\onecolumngrid
\appendix
\section{Escaped MCP Trace JSON Structure}
\label{app:trace_json}

\begin{verbatim}
{
  "question_id": "clin-real-v2-shuf-003",
  "user_question": "Summarize beta blockers in heart failure with atrial fibrillation.",
  "category": "literature_review",
  "route": "pubmed_research",
  "tool_calls": [
    "search_pubmed_key_words",
    "get_pubmed_article_metadata"
  ],
  "full_tool_outputs": [
    {
      "tool_name": "get_pubmed_article_metadata",
      "source_id": "tool_output::get_pubmed_article_metadata",
      "source_role": "complete_tool_output",
      "chunk_count": 3,
      "text": "Complete tool output for get_pubmed_article_metadata.\\n[1] ..."
    },
    {
      "tool_name": "search_pubmed_key_words",
      "source_id": "tool_output::search_pubmed_key_words",
      "source_role": "complete_tool_output",
      "chunk_count": 12,
      "text": "Complete tool output for search_pubmed_key_words.\\n[1] ..."
    }
  ],
  "original_evidence_chunk_count": 15,
  "assistant_answer": "A source-grounded answer summarizing the literature...",
  "final_reply_to_user": "A source-grounded answer summarizing the literature...",
  "elapsed_s": 19.97
}
\end{verbatim}

\section{Random-Forest Calibration Input and Output}
\label{app:calibration_json}

\begin{verbatim}
{
  "claim_id": "clin-real-v2-shuf-014::c01",
  "candidate_claim": "The patient is Carla Mendoza.",
  "routed_source": {
    "source_id": "tool_output::load_patient_history",
    "tool_name": "load_patient_history",
    "route_score": 0.224,
    "routing_margin": 0.224
  },
  "calibrator_input": {
    "categorical_features": {
      "verifier_label": "entailment",
      "category": "chart_plus_literature",
      "pred_tool": "load_patient_history",
      "status": "routed_entailment"
    },
    "numeric_features": {
      "route_score": 0.224,
      "pred_score": 0.927,
      "routing_margin": 0.224,
      "term_overlap": 1.000,
      "all_term_overlap": 1.000,
      "claim_terms_n": 3,
      "protected_n": 1,
      "protected_missing": 0,
      "protected_missing_all": 0,
      "evidence_chunks_n": 4,
      "claim_len": 5,
      "has_stated_attr": 0,
      "score_x_route": 0.208,
      "score_x_overlap": 0.927,
      "route_x_overlap": 0.224,
      "neutral": 0,
      "entailment": 1,
      "contradiction": 0,
      "weak_route_high_score": 0,
      "strong_route": 0,
      "score_x_route_squared": 0.043
    }
  },
  "calibrator": {
    "model": "RandomForestClassifier",
    "n_estimators": 400,
    "max_depth": 5,
    "min_samples_leaf": 8,
    "class_weight": "balanced",
    "threshold": 0.650
  },
  "calibrator_output": {
    "support_probability": 0.9996,
    "support_decision": "supported",
    "pred_relation": "entailment",
    "pred_source_id": "tool_output::load_patient_history",
    "pred_tool_id": "load_patient_history"
  },
  "downstream_use": {
    "support_passes_to_attribution_matching": true,
    "answer_level_policy": "block if any claim fails support or attribution"
  }
}
\end{verbatim}

\section{Escaped Claim, Routing, NLI, and Repair Example}
\label{app:example_json}

\begin{verbatim}
{
  "trace_evidence": {
    "tool_name": "load_patient_history",
    "source_id": "tool_output::load_patient_history",
    "text": "id: bench-6; name: Felipe Costa; label: Asthma; status: active."
  },
  "atomic_claim": {
    "text": "Felipe Costa (bench-6) has active condition Asthma",
    "stated_attribution": "According to PubMed evidence."
  },
  "routing": {
    "routed_source_id": "tool_output::load_patient_history",
    "tool_name": "load_patient_history",
    "score": 0.637,
    "margin": 0.413
  },
  "nli": {
    "relation": "entailment",
    "protected_values": ["bench-6"],
    "unsupported_content_token_indices": []
  },
  "attribution": {
    "status": "conflation",
    "reason": "Supported by patient-history evidence, not PubMed evidence."
  },
  "rarr": {
    "status": "corrected",
    "replacement": "FHIR history: Felipe Costa (bench-6) has active Asthma."
  }
}
\end{verbatim}

\end{document}